\documentclass[10pt,twocolumn,letterpaper]{article}

\usepackage{cvpr}              
\usepackage{array}
\usepackage[table]{xcolor}  
\usepackage{multirow}
\usepackage{bm}
\usepackage{arydshln}
\usepackage{graphicx}
\usepackage{caption}
\usepackage{caption}
\usepackage{tcolorbox}
\usepackage{etoolbox}
\usepackage{hhline} 
\usepackage{pdfpages}

\newcommand{\ourMthd}{OmniRefiner}

\renewcommand{\eqref}[1]{Eqn.~(\ref{#1})}

\definecolor{cvprblue}{rgb}{0.21,0.49,0.74}
\usepackage[pagebackref,breaklinks,colorlinks,allcolors=cvprblue]{hyperref}

\def\paperID{1918} 
\def\confName{CVPR}
\def\confYear{2026}

\title{OmniRefiner: Reinforcement-Guided Local Diffusion Refinement}

\author{
  Yaoli Liu\textsuperscript{1,4} \quad
  Ziheng Ouyang\textsuperscript{2} \quad
  Shengtao Lou\textsuperscript{4} \quad
  Yiren Song\textsuperscript{3, 4 $\dagger$} \\
  \textsuperscript{1} Zhejiang University, \textsuperscript{2} Nankai University, \textsuperscript{3} National University of Singapore, \textsuperscript{4} Creatly.ai 
}

\begin{document}

\twocolumn[{%
\renewcommand\twocolumn[1][]{#1}%
\maketitle
\begin{center}
   \captionsetup{type=figure}
    \vspace{-0.5cm}    
\includegraphics[width=\linewidth]{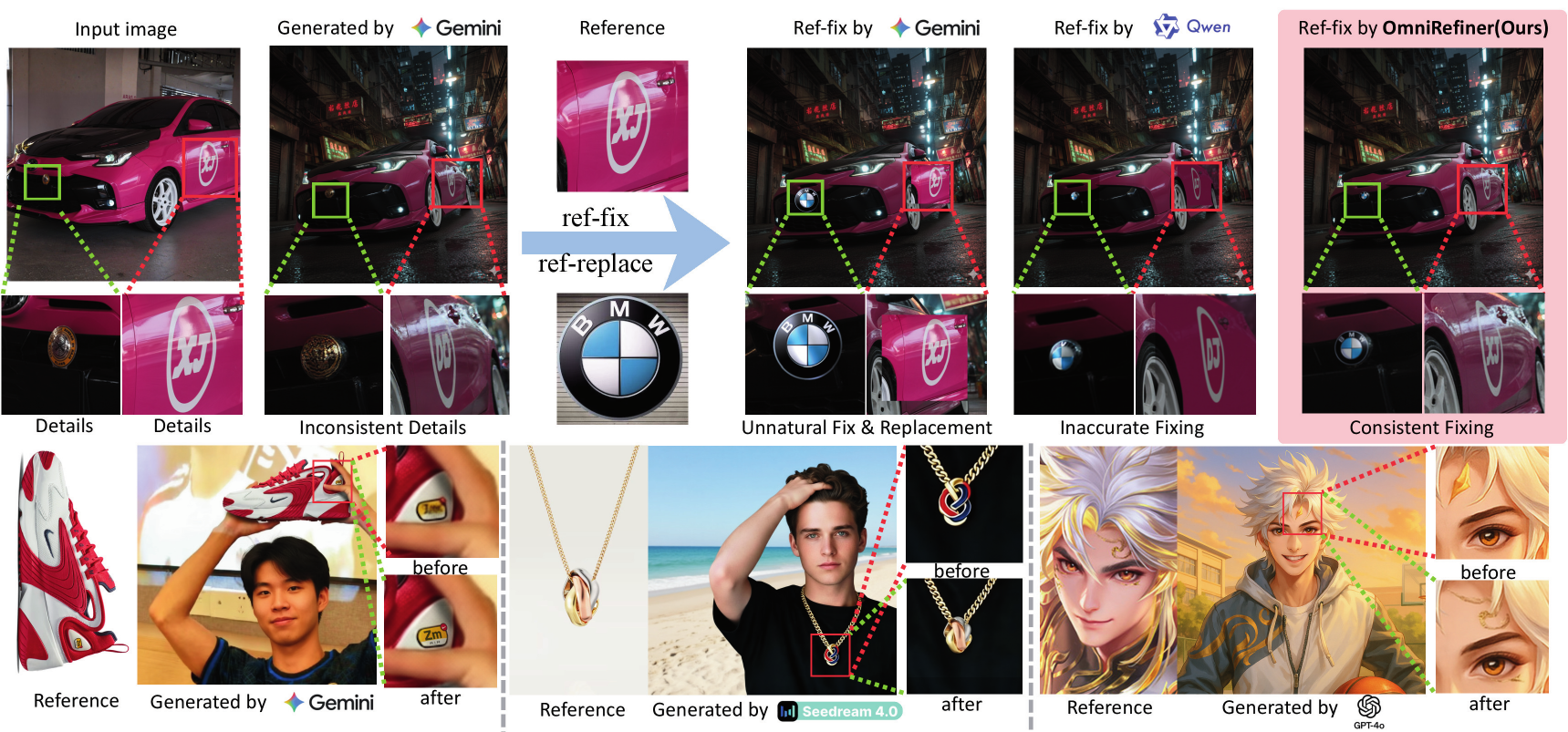}    
    \setlength{\abovecaptionskip}{0pt}
    \caption{We propose \textbf{\ourMthd{}}, a model capable of refining image details based on reference images. It can accurately restore various fine details such as logos, text, facial features, and intricate patterns, showing great potential for downstream applications in e-commerce, facial beautification, and advertising.
}
    \label{teaser} 
    \vspace{5pt}
\end{center}   
}]

\begingroup
\renewcommand\thefootnote{}
\footnotetext{$\dagger$ Corresponding author.}
\endgroup

\begin{abstract}

Reference-guided image generation has progressed rapidly, yet current diffusion models still struggle to preserve fine-grained visual details when refining a generated image using a reference. This limitation arises because VAE-based latent compression inherently discards subtle texture information, causing identity- and attribute-specific cues to vanish. Moreover, post-editing approaches that amplify local details based on existing methods often produce results inconsistent with the original image in terms of lighting, texture, or shape. To address this, we introduce \ourMthd{}, a detail-aware refinement framework that performs two consecutive stages of reference-driven correction to enhance pixel-level consistency. We first adapt a single-image diffusion editor by fine-tuning it to jointly ingest the draft image and the reference image, enabling globally coherent refinement while maintaining structural fidelity. We then apply reinforcement learning to further strengthen localized editing capability, explicitly optimizing for detail accuracy and semantic consistency. Extensive experiments demonstrate that \ourMthd{} significantly improves reference alignment and fine-grained detail preservation, producing faithful and visually coherent edits that surpass both open-source and commercial models on challenging reference-guided restoration benchmarks. Our project homepage is available at
\href{https://github.com/yaoliliu/OmniRefiner}{https://github.com/yaoliliu/OmniRefiner}
\end{abstract}

\section{Introduction}
\label{sec:intro}
Recently, image editing models have advanced rapidly. Starting from text-to-image models that perform conditional editing through training modules such as ControlNet~\cite{zhang2023adding}, to current specialized editing and generation models~\cite{nanobanana, wu2025qwen, labs2025flux1kontextflowmatching, huang2025arteditor} trained on large-scale editing datasets, model capabilities have significantly improved across various image editing tasks, including virtual try-on, multi-image fusion, face replacement, and style transfer. Despite this progress, even state-of-the-art diffusion models struggle to preserve fine-grained identity and structural fidelity---notably for logos, text, facial micro-geometry, and texture-critical regions. A major culprit is the aggressive compression in latent diffusion pipelines, where VAE encoders inevitably discard subtle local cues. 
As a result, when users expect precise transfer from a reference crop, models often over-smooth or distort details, degrading reference fidelity.

To address above issues, we introduce \textbf{\ourMthd{}}, a universal post-refinement module for reference detail-consistent enhancement.
Specifically, given a generated image and a reference patch, \ourMthd{} operates on zoomed local regions in reference patch to restore high-frequency details in generated image while preserving global consistency including light and background. 
However, this task presents three challenges:
(1) the refined region must align with the reference under perspective, lighting, and geometric variation rather than naive copy-paste;
(2) non-edited areas must remain strictly identical to the original image to avoid artifacts upon reintegration;
(3) the method must generalize across object categories, scene types, and generator models (open-source and commercial).

These challenges motivate three design principles in \ourMthd{}. 
\emph{For (1)}, we adopt 
FLUX.1-Kontext-dev, a single-image editing transformer, into a dual-input conditional generator and use bidirectional attention between target and reference tokens, enabling precise, content aware detail transfer under spatial variations. 
\emph{For (2)}, we introduce a supervised fine-tuning (SFT) stage with explicit locality awareness: the model learns to edit only the masked region while preserving the remainder verbatim, thereby preventing collateral changes. 
\emph{For (3)}, we construct a \emph{large-scale synthetic triplet pipeline} that automatically produces diverse training tuples via image editing and VLM-guided cropping, covering rich categories, materials, and degradations to ensure strong cross-domain and cross-backbone generalization.

While the above addresses spatial alignment, locality, and generalization, we further observe that micro-textures such as thin text strokes, serial numbers and fabric weaves can remain under-fit due to diffusion smoothing and supervision imbalance. To enhance \emph{detail consistency}, we introduce a second stage training strategy based on GRPO: patch-wise rewards combine a perceptual metric DreamSim with a masked pixel term, selectively sharpening high-frequency regions without perturbing the background, which serves as a precision tuner atop SFT, improving robustness to illumination and geometric changes and stabilizing fine-detail reconstruction.

To support training and evaluation, we curate a 30K-triplet benchmark of degraded targets, clean references, and ground truth outputs built by our synthetic pipeline. The dataset enables scalable supervision for SFT and reliable reward computation for RL.

Our main contributions are summarized as follows:
\begin{itemize}
\item 
We propose \textbf{\ourMthd{}}, a universal reference-guided detail correction module that enhances diffusion outputs without disturbing global structure. 

\item 
We introduce a two-stage refinement paradigm: dual-input in-context SFT for alignment/locality and position embedding extension, followed by GRPO-based patch rewards to boost fine-detail consistency. 

\item  
We build a 30K localized refinement dataset via an automated four-stage data collecting and creating pipeline based on image-editing model and VLM. Experiments demonstrate our model has state-of-the-art fidelity across diverse content and generator backbones.
\end{itemize}

\section{Related work}

\subsection{Diffusion Models}

Diffusion models have emerged as a powerful generative paradigm for producing high-fidelity images through iterative denoising. The introduction of DDPM~\cite{ho2020denoising}, subsequent advances such as Latent Diffusion Models~\cite{song2020denoising} and Latent consistency models~\cite{luo2023latent} have enhanced its usability. Recent years, DiT~\cite{peebles2023scalable} and Flow matching~\cite{lipman2022flow} have significantly improved efficiency and scalability by operating in compressed latent spaces and replacing U-Net~\cite{ronneberger2015u} backbones with Transformer-based architectures. Open-source text-to-image models have evolved from primarily UNet-based models, such as Stable Diffusion~\cite{rombach2022high} and Stable Diffusion XL~\cite{podell2023sdxl}, to increasingly DiT-based models, including FLUX~\cite{flux2024}, Stable Diffusion 3.5~\cite{esser2024scaling}, and Qwen-Image~\cite{wu2025qwen}. For a period of time, reference-conditioned generation was typically achieved by training ControlNet~\cite{zhang2023adding} on top of existing text-to-image models \cite{ma2024followyouremoji, ma2025controllable, ma2025followcreation, ma2025followfaster, ma2025followyourclick, ma2025followyourmotion, song2024processpainter, zhang2024stable, zhang2025stable, chen2025transanimate}, or by introducing an encoder~\cite{Li_2024_CVPR, Chen_2024_CVPR, Cao_2023_ICCV, mao2025ace++, mou2024t2i} capable of referencing an image to inject specific features into the latent space of the text-to-image model. However, large-scale image editing models are now becoming the mainstream.  With the architectural transition to Diffusion Transformers~\cite{dit}, recent approaches such as EasyControl~\cite{zhang2025easycontrol} have achieved image-conditioned generation within MM-DiT frameworks and inspired subsequent works ~\cite{song2025omniconsistency, gong2025relationadapter, jiang2025personalized, wang2025diffdecompose, song2025makeanything, song2025layertracer, guo2025any2anytryon,  lu2025easytext, shi2025wordcon, shi2024fonts}.

\subsection{Image Editing Models}

\begin{figure*}[tbp]
  \centering
  \includegraphics[width=1\textwidth]{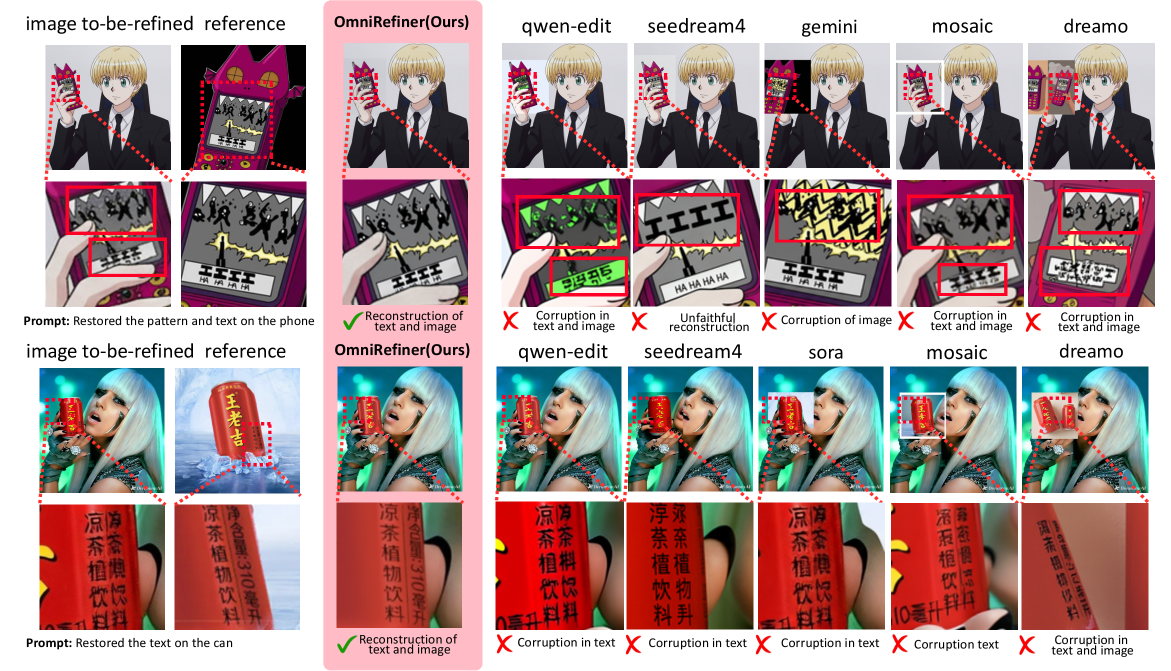}
  \caption{Compared with the state-of-the-art multi-image editing methods, our approach achieves not only faithful reconstruction of the original image in reference–repair tasks, but also excellent performance in various reconstruction scenarios including text, patterns, facial details, and object details. In contrast, existing methods often fail to remain faithful to the original image during repair or are unable to recover text and fine details.}
  \label{fig:method_comp}
\end{figure*}

Although research on training-free or post-training text-to-image editing models is still ongoing~\cite{Kulikov2024FlowEditIT, Brack2023LEDITSLI, Deutch2024TurboEditTI, Hertz2022PrompttoPromptIE, mou2025dreamo, she2025mosaicmultisubjectpersonalizedgeneration,xiao2025omnigen, chen2025xverseconsistentmultisubjectcontrol}, recent large-scale image editing models based on the DiT architecture, such as Bagel~\cite{deng2025emerging}, FLUX.1-Kontext~\cite{labs2025flux1kontextflowmatching}, and Qwen-Image-Edit~\cite{wu2025qwen}, have demonstrated capabilities that far surpass previous approaches, as researchers continue to push the limits of diffusion models under the scaling law. Both open-source and closed-source editing models now achieve impressive performance in large-scale transformations such as object composition, action modification, and viewpoint changes. However, their generated subjects still suffer from deficiencies in fine textures, facial details, and textual elements. Moreover, while current models exhibit strong text-guided editing abilities, they remain incapable of accurately performing detail restoration when users expect the model to reference a given image. In such cases, these models either fail to edit at all or produce incorrect reference-based modifications, as shown in Fig.~\ref{fig:method_comp}

\subsection{Reinforcement Learning in Image Generation}

Reinforcement learning (RL) has recently emerged as a promising paradigm for improving generative models, particularly in aligning generation with human preferences and fine-grained constraints. Works such as RLHF for text-to-image diffusion~\cite{Yang2023UsingHF, wallace2024diffusion, Black2023TrainingDM} demonstrate that reward-driven optimization can enhance visual alignment, aesthetics, and user satisfaction. With the growing popularity of the GRPO~\cite{shao2024deepseekmath} algorithm in large language models (LLMs), an increasing number of researchers have begun exploring its application in flow-matching~\cite{lipman2022flow} models to further enhance reinforcement learning performance, as seen in works such as DanceGRPO~\cite{xue2025dancegrpo} and FlowGRPO~\cite{liu2025flow}.
However, current applications of reinforcement learning in diffusion models primarily focus on aligning overall generation results with human preferences. In contrast, our detail restoration task requires the model to pay closer attention to local details. Inspired by ~\cite{kotar2023these}, we design a reward function specifically tailored to emphasize local fine-grained features.

\section{Method}
\label{sec:method}

\begin{figure*}[tbp]
  \centering
  \includegraphics[width=1\textwidth]{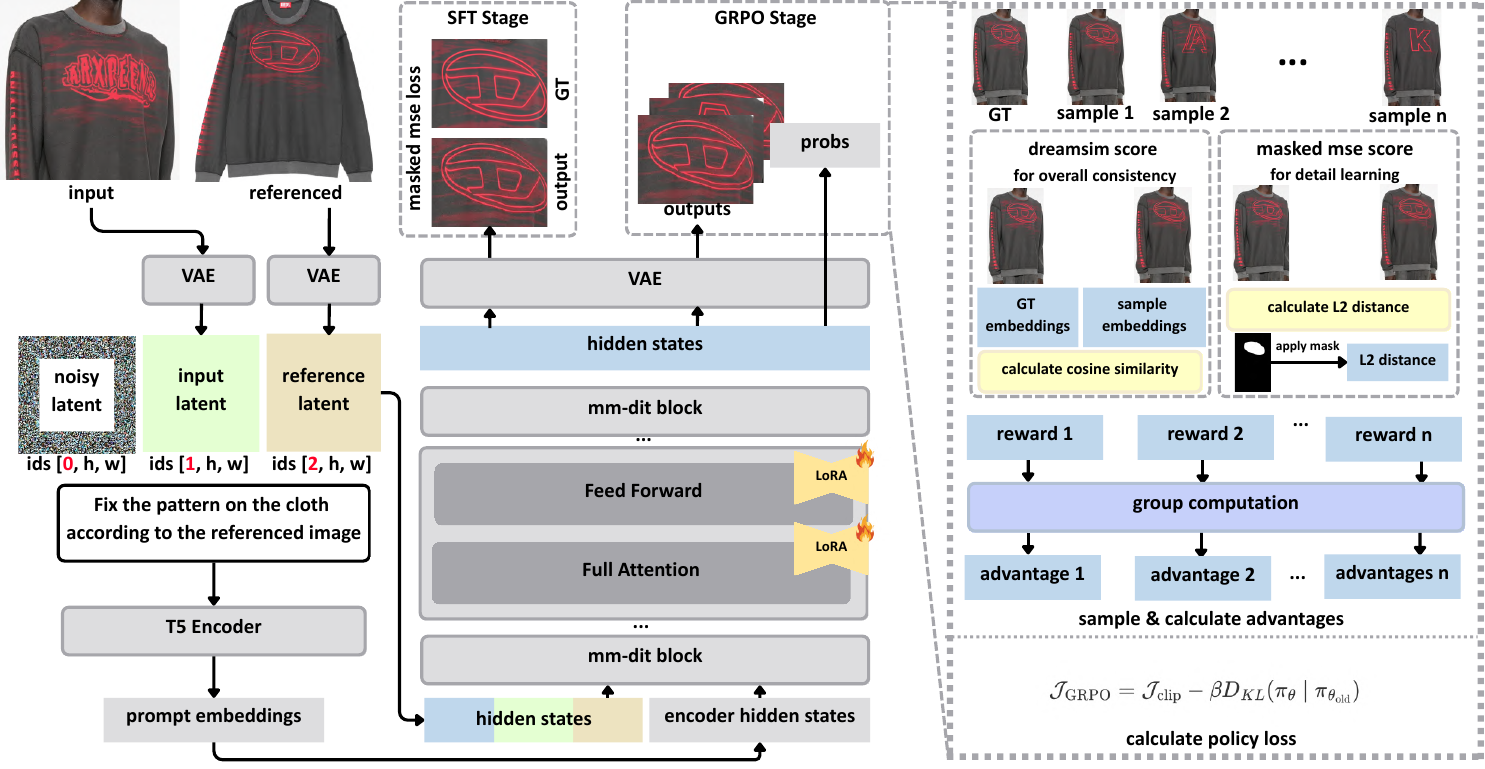}
  \caption{Overall architecture of OmniRefiner. Our framework adopts a two-stage training pipeline. In the first stage, we perform supervised fine-tuning (SFT) to enable dual-input detail restoration while preserving global structure. In the second stage, we apply GRPO-based reinforcement learning to further enhance fine-grained consistency and local repair quality. This joint design enables precise reference-guided refinement with high visual fidelity.}
  \label{fig:method_arch}
\end{figure*}

In Sec.~\ref{sec:overall}, we outline the overall two-stage refinement framework.
In Sec.~\ref{sec:sft}, we introduce the supervised dual-input diffusion architecture for localized detail restoration.
In Sec.~\ref{sec:rl}, we present a GRPO-based reinforcement learning objective to further enhance fine-grained consistency.
In Sec.~\ref{sec:data}, we describe our automated synthetic triplet data pipeline for scalable training and strong generalization.

\subsection{Overall Architecture}
\label{sec:overall}

We present \textbf{OmniRefiner}, a two-stage framework for reference-guided detail refinement. Given a to-be-refined image $I \in \mathbb{R}^{H\times W\times 3}$, reference crop $R \in \mathbb{R}^{h\times w\times 3}$ and an edit instruction $P$, our goal is to produce a refined image $\hat{I}$ such that local details in the refinement region $\Omega \subseteq \{1,\dots,H\}\times\{1,\dots,W\}$ match those in $R$ while preserving $I$ outside $\Omega$:
\begin{equation}
\hat{I}_\Omega \approx R_\Omega, 
\qquad 
\hat{I}_{\bar{\Omega}} = I_{\bar{\Omega}},
\end{equation}
where $\bar{\Omega}$ denotes the complement of $\Omega$ and the equality outside $\Omega$ is enforced up to numerical tolerance.
Our two-stage pipeline is consist of SFT stage and RL stage.
In SFT stage, we adapt FLUX.1-Kontext-dev as our base model, and transform it into a dual-input conditional generator that receives both $I$ and $R$. It learns to preserve global semantics from $I$ while selectively injecting high-frequency details from $R$. While in RL stage, We further optimize the model with patch-wise reward and dreamsim reward to further improve robustness against perspective, illumination, and geometric variations. Our overall architecture is illustrated in Fig.~\ref{fig:method_arch}.  
                    
\subsection{Supervised Finetuning for Basic Understanding}
\label{sec:sft}
We attempt to perform local refinement using current state-of-the-art multi-input models; however, as shown in Fig.~\ref{fig:method_comp}, they suffer from copy-paste or inconsistent problems such as visible seams, color bleeding, and structural drift, failing to remain consistent with the global content of $I$.
To address the above issues, we adopt SFT training, trying to make the model learn where and how to integrate reference details while respecting the global composition while extending the model to accept two images(input and reference) as input. Specifically, we employ Bidirectional attention and local mask loss so that detail transfer is context-aware instead of being a rigid local paste.

\noindent \textbf{Bidirectional Attention.}
In our approach, we employ a bidirectional attention mechanism, which allows the model to attend to the noisy latent, prompt, input latent, and reference latent simultaneously. Specifically, the model first encodes the image \( I \) and the reference \( R \) into their respective latent representations, \( c_I \) and \( c_R \). After applying position encoding cloning, the latent tokens are concatenated along the sequence dimension to perform joint attention. The attention mechanism is formulated as follows:

\vspace{-10pt}
\begin{equation}
\text{MMA}([z; c_I; c_R; c_T]) = \text{softmax}\left(\frac{QK^T}{\sqrt{d}}\right)V,
\end{equation}
where $[z; c_I; c_R; c_T]$ denotes the concatenation of the noised latent tokens $z$, the image condition tokens $c_I$, the reference condition tokens $c_R$, and the prompt tokens $c_T$, allowing the conditional and denoising branches to interact as needed. Here, $Q$, $K$, and $V$ represent the query, key, and value matrices, respectively, which are derived from the concatenated input via linear projections. The term $d$ is the dimension of the key features, serving as a scaling factor $\frac{1}{\sqrt{d}}$ to ensure gradient stability.

In this process, we apply position encoding (PE) for each latent. Specifically: The noisy latent \( z \) uses position encoding \( \text{ids}[0, h, w] \), where \( h \) and \( w \) represent the height and width of the image grid.
The input latent \( c_I \) and reference latent \( c_R \), use position encoding \( \text{ids}[1, h, w] \) and  \( \text{ids}[2, h, w] \), respectively.
This approach enables the model to maintain spatial consistency while performing precise detail transfer and denoising. The use of bidirectional attention enhances the model's ability to process both local and global structures, leading to improved image editing results.

\noindent \textbf{Weighted Mask Loss.}
Follow the setting adopted by~\cite{lin2025uniworld}, we define a \textbf{weighted mask loss} to ensure that the model focuses refinements on the desired region $\Omega$ while preserving the background $\bar{\Omega}$. This approach computes the error across the entire image but applies different weights to the target and background regions.

Specifically, we define a binary mask $M \in \{0,1\}^{H \times W}$, where $M(p) = 0$ for pixels $p \in \Omega$ (the target region) and $M(p) = 1$ for pixels $p \notin \Omega$ (the background).
From this mask, we derive a pixel-wise weight matrix $W$. The objective is to \textbf{up-weight} the loss within the target region $\Omega$ to emphasize refinement, while maintaining a standard weight of 1 for the background $\bar{\Omega}$ to penalize unwanted changes.
The weight $W(p)$ for each pixel $p$ is defined as:
\begin{equation}
W(p) =
\begin{cases}
\dfrac{H \times W}{\sum_{p} (1 - M(p))}, & \text{if } M(p) = 0, \ p \in \Omega, \\[8pt]
1, & \text{if } M(p) = 1, \text{ else.}
\end{cases}
\end{equation}
The weighted mask loss is then computed as the mean of the weighted pixel-wise squared errors:
\begin{equation}
\label{eq:sft-loss}
\mathcal{L}_{\mathrm{mask}} = \frac{1}{H \times W} \sum_{p} W(p) \left( \hat{I}(p) - I^\star(p) \right)^2,
\end{equation}
where $\hat{I}(p)$ is the predicted pixel value at position $p$, and $I^\star(p)$ is the corresponding ground truth pixel value.

This loss function strongly encourages the model to minimize the error within the region of interest $\Omega$, simultaneously, the standard-weighted term for $\bar{\Omega}$ ensures the background structure is preserved.

\subsection{Reinforcement Learning for Enhanced Ability}
\label{sec:rl}

Although SFT equips the model with a strong alignment prior, certain micro details (e.g., text edges, serial numbers, micro textures) are easily underfit because $\mathcal{L}_{\mathrm{SFT}}$ is dominated by global denoising statistics. We therefore adopt \emph{reward-driven} optimization to explicitly push the model toward \emph{patch-level} perceptual similarity and pixel accuracy within $\Omega$, while leaving $\bar{\Omega}$ intact. We split $\Omega$ into a non-overlapping patch set $\mathcal{P}_\Omega=\{P_k\}_{k=1}^{K}$ (e.g., $512{\times}512$ windows). Let $\hat{I}[P_k]$ denote the cropped prediction on patch $P_k$, and $I^\star[P_k]$ the corresponding ground truth. 

\noindent \textbf{GRPO objective.}
Given a prompt \(\bm{c}\), the flow model \(p_{\theta}\) generates a batch of \(G\) images \(\{\bm{x}^i_0\}_{i=1}^G\) along with their reverse-time trajectories \(\{(\bm{x}^i_T, \bm{x}^i_{T -1}, \cdots, \bm{x}^i_0)\}_{i=1}^G\). GRPO optimizes the policy model through the following objective:
\begin{equation}
\begin{aligned}
\label{eq:grpo}
\mathcal{J}_{\mathrm{GRPO}}(\theta)
&= \mathbb{E}_{\mathbf{c}\sim\mathcal{C},\,\{\mathbf{x}^i\}\sim\pi_{\theta_{\mathrm{old}}}(\cdot\mid\mathbf{c})}
\\[-3pt]
&\!\!\!\!\!\!\!\!\!\!\!\!\!\!\!\!\!\!\!\!\!
\Bigg[
\frac{1}{G}\sum_{i=1}^{G}\,\frac{1}{T}\sum_{t=0}^{T-1}
\Big(
\min\!\big(r_t^i(\theta)\hat{A}_t^i,\,
\operatorname{clip}(r_t^i(\theta),1-\varepsilon,1+\varepsilon)\hat{A}_t^i\big)
\Big)
\\[-5pt]
&\qquad
-\beta\, D_{\mathrm{KL}}\!\left(\pi_{\theta}(\cdot\mid\mathbf{c})\,\Vert\,\pi_{\mathrm{ref}}(\cdot\mid\mathbf{c})\right)
\Bigg],
\end{aligned}
\end{equation}
where 
$$r^i_t(\theta) =\frac{p_{\theta}(\bm{x}^i_{t-1} \mid \bm{x}^i_t, \bm{c})}{p_{\theta_{\text{old}}}(\bm{x}^i_{t-1} \mid \bm{x}^i_t, \bm{c})}.$$
The advantage term \(\hat{A}^i_t\) is obtained by standardizing rewards across the batch:
$$\hat{A}^i_t = \frac{R(\bm{x}^i_0, \bm{c}) - \text{mean}(\{R(\bm{x}^j_0, \bm{c})\}_{j=1}^G)}{\text{std}(\{R(\bm{x}^j_0, \bm{c})\}_{j=1}^G)}.$$

\begin{figure*}[h]
  \centering
  \includegraphics[width=1\textwidth]{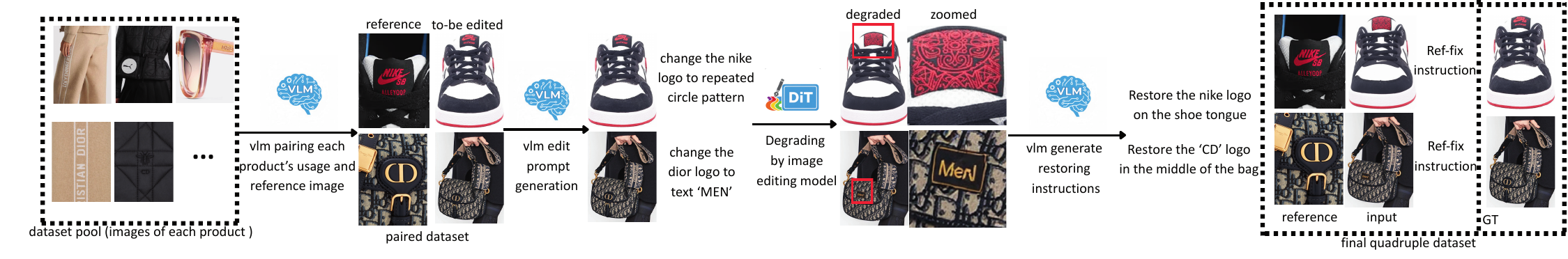}
  \caption{We adopt a four-stage data pipeline. First, a VLM pairs images of the same product with consistent styles and reasonable viewpoints. Second, it generates fine-grained editing instructions for one image in each pair. Third, an image editing model executes these edits using the pre-edit image as ground truth, forming our (input, reference, ground truth) triplet dataset.Finally, the VLM generates an instruction guiding the model to restore the input using the reference, based on the input, reference, and ground truth.}
  \label{fig:data_pipe}
\end{figure*}

\noindent \textbf{Mask Pixel Reward.}
We complement the perceptual term with a masked MSE:
\begin{equation}
\label{eq:rmse}
\begin{aligned}
\mathcal{R}_{\mathrm{mm}}
&=
-
\frac{1}{|\Omega|}
\sum_{p\in \Omega}
\big(
\hat{I}(p) - I^\star(p)
\big)^2.
\end{aligned}
\end{equation}

\noindent \textbf{Perceptual Reward.}
Let $f(\cdot)$ be a frozen DreamSim embedding. We compute a perceptual similarity reward per patch and then average:
\begin{equation}
\label{eq:rsim}
\mathcal{R}_{\mathrm{ds}}
=
-\frac{1}{K}
\sum_{k=1}^{K}
\left\|
f\!\left(\hat{I}[P_k]\right)
-
f\!\left(I^\star[P_k]\right)
\right\|_2.
\end{equation}
The overall scalar reward is calculated as:
\begin{equation}
\label{eq:r-total}
\mathcal{R}
=
(1-\lambda)\mathcal{R}_{\mathrm{ds}}
+
\lambda\, \mathcal{R}_{\mathrm{mm}},
\end{equation}
where $\lambda>0$ balances overall perceptual similarity and detail pixel-wise accuracy.

\noindent \textbf{ODE to SDE GRPO.}
Following~\cite{liu2025flow}, we apply GRPO to the flow-matching model using the ODE-to-SDE formulation. 
Specifically, for a set of inputs $\big(I, R, I^\star, P\big)$ and an exploration count of $m$, 
we sample $m$ trajectories. During each sampling step, a certain amount of random noise is added 
to encourage stochastic exploration along the path. For a trajectory with exploration probability $p$, 
we predict it using
\vspace{-10pt}
\begin{equation}
P(x_{\text{p}}) = \frac{1}{\sigma_{\text{s}} \sqrt{2\pi}} 
\exp\left( -\frac{(x_{\text{p}} - \mu_{\text{p}})^2}{2 \sigma_{\text{s}}^2} \right),
\end{equation}
where $x_{\text{p}}$ is the predicted sample at current timestep, $\sigma_{\text{s}}$ is a time-dependent parameter, and $\mu_{\text{p}}$ 
is the mean of the Gaussian distribution predicted for the current denoising step. 

After obtaining rewards, we compute the advantage for each trajectory and optimize the GRPO objective \eqref{eq:grpo} 
via gradient descent.
\subsection{Automated Dataset Pipeline}
\label{sec:data}

To train at scale, we construct quadruple $\big(I, R, I^\star, P\big)$ automatically. Starting from clean images $I^\star$, we first sample a region $\Omega$ using a VLM-based saliency/objectness selector, then produce a degraded variant $I=\mathrm{Degrade}(I^\star,\Omega)$ (blur, compression, downsampling, text/logo erosion, lighting shifts, color/texture change, text remove/change), and obtain $R=\mathrm{Crop}(I^\star,\Omega)$ as the reference. This yields large, diverse supervision for ~\eqref{eq:sft-loss} and provides reliable targets for the patch-wise RL in ~\eqref{eq:grpo} -- \eqref{eq:rsim}.

As illustrated in Fig.~\ref{fig:data_pipe}, we adopt a three-stage data processing pipeline. Our raw data are collected and categorized based on individual products; however, each product may still contain multiple styles, colors, or large variations in viewing angles (e.g., top-down views of shoes or sole images). Therefore, in the first stage, we employ a Vision-Language Model (VLM) to pair images of the same product that share consistent styles and have reasonable viewpoint variations. In the second stage, the VLM generates fine-grained editing instructions for one image in each pair. In the third stage, we apply an image editing model to perform the instructed edits, using the pre-edit image as the ground truth. In the fourth stage, we provide the input, reference, and ground truth to the VLM and ask it to generate an instruction that edits the input into the ground truth based on the reference. This process yields our quadruple dataset.

\section{Experiment.}

\subsection{Experiment Setup.}

\textbf{Experiment Details.}
\begin{figure}[tbp]
  \centering
  \begin{subfigure}[b]{0.23\textwidth}
    \centering
    \includegraphics[width=\linewidth]{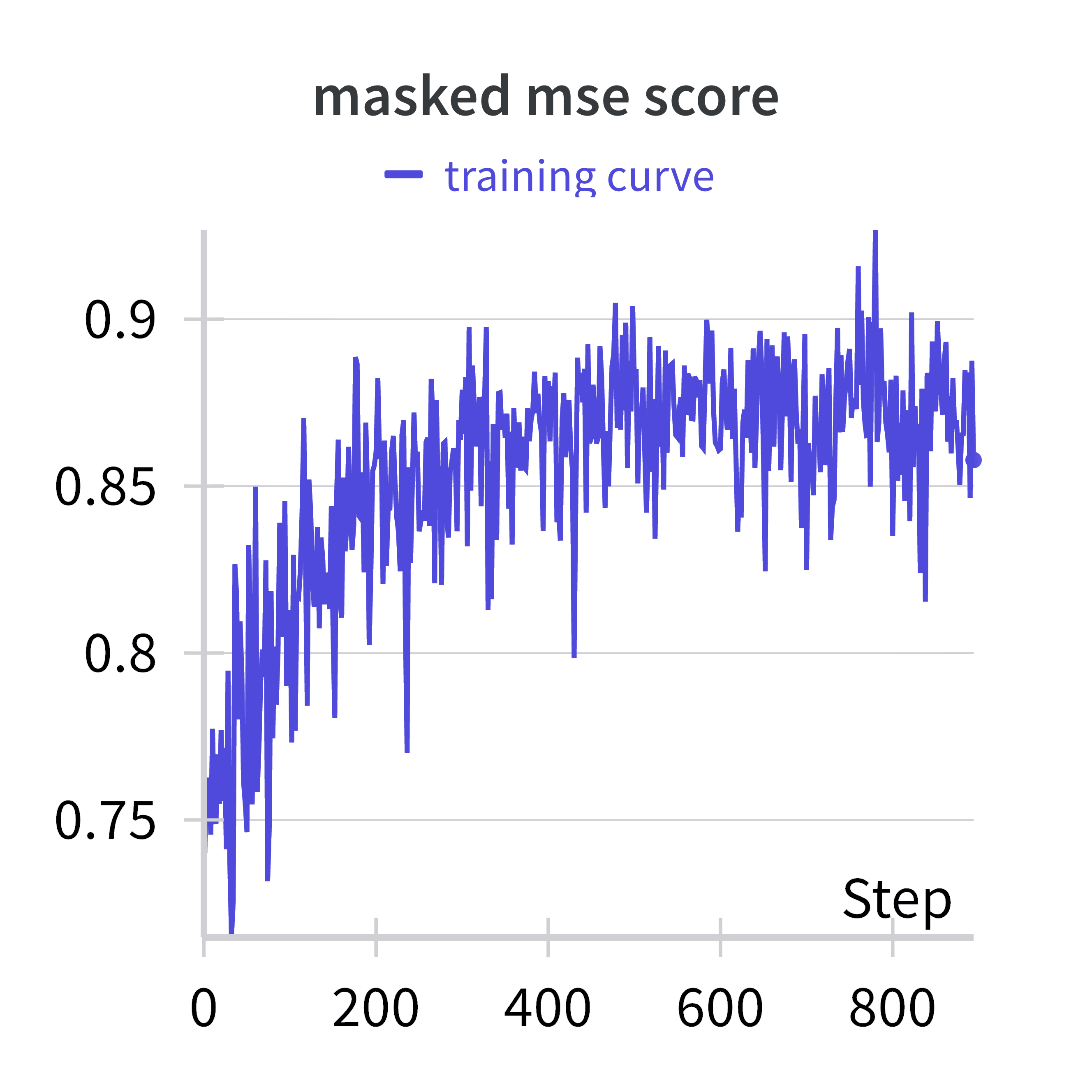}
  \end{subfigure}
  \hfill 
  \begin{subfigure}[b]{0.23\textwidth}
    \centering
    \includegraphics[width=\linewidth]{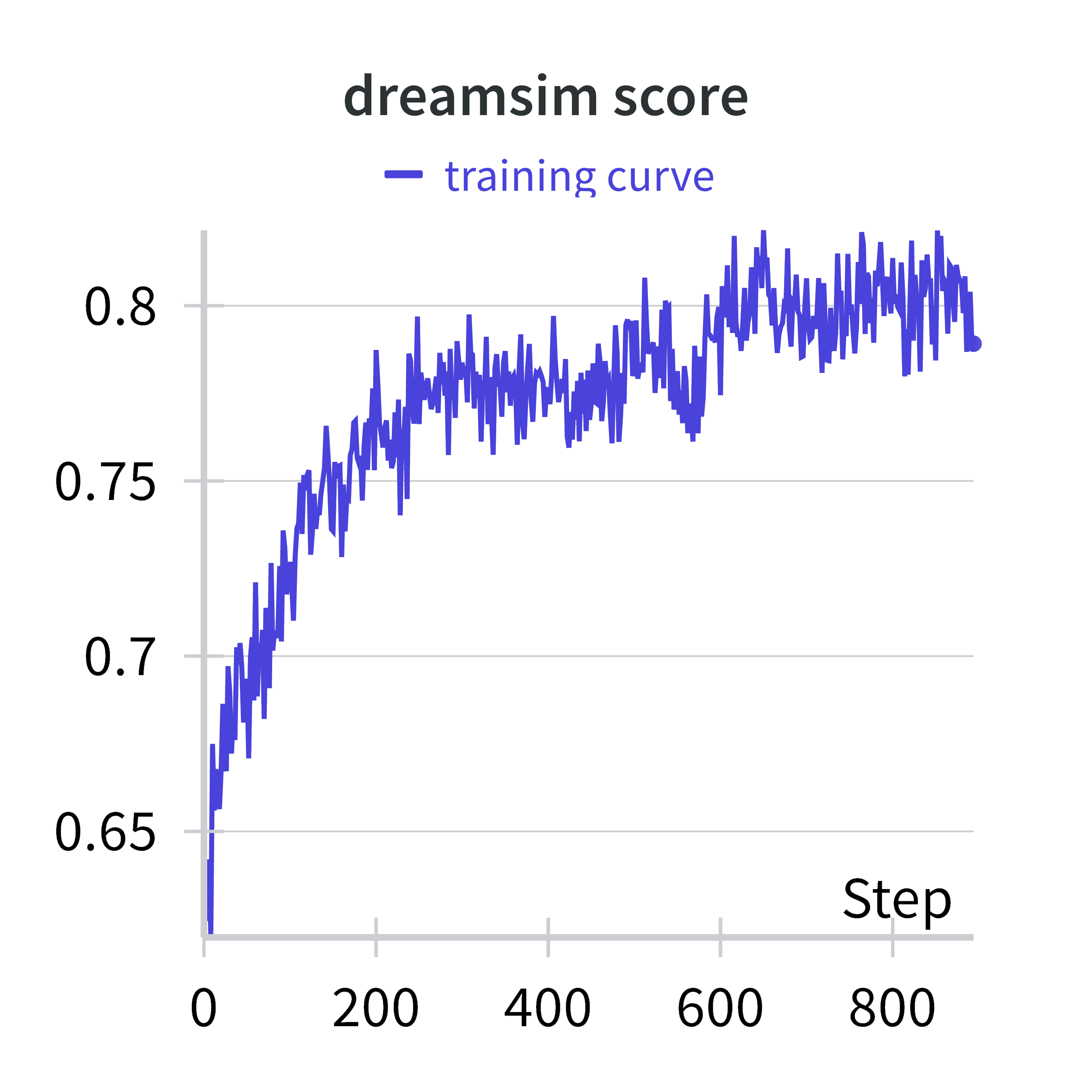}
  \end{subfigure}
  
  \caption{The DreamSim reward curve and the masked MSE reward curve demonstrate the process of how our model aligns with the reward functions during GRPO.}
  \vspace{-10pt}
  \label{fig:ablation_comparison}
\end{figure}

\begin{table*}[t]\small
\centering
\scriptsize{
\captionsetup{font=small}
\caption{{
Compared to existing SOTA methods, our approach demonstrates improvements across various image similarity metrics, face similarity metrics for face restoration, and evaluations by advanced VLMs.
}}
\label{tab:compare_sota_vqa}
\vspace{-10pt}
\resizebox{\linewidth}{!}{
\setlength\tabcolsep{3.pt}
\renewcommand\arraystretch{1.1}
\begin{tabular}{r||cccccc|cc|c}
\hline
\hline
\rowcolor{gray!20}
 & 
\multicolumn{6}{c|}{\textbf{Overall Local Refine}} & 
\multicolumn{2}{c|}{\textbf{Face Local Refine}} &
\multicolumn{1}{c}{\textbf{Text Local Refine}}
\\
\cline{2-10} 
\rowcolor{gray!20}
\multirow{-2}{*}{Methods}  
& Dreamsim$\downarrow$ & DINOv2$\uparrow$ & DINOv3$\uparrow$ & Clip-I$\uparrow$ & MSE$\downarrow$ & VLM Score$\uparrow$
& ArcFace$\uparrow$ & LVFace$\uparrow$
& VLM Text Identify$\uparrow$
\\
\hline\hline
DreamO 
& 0.3475 & 0.5005 & 0.4335 & 0.8861 & 0.1506 & 28.04
& 0.5082 & 0.3446  
& 2.874
\\
\rowcolor{gray!10} Mosaic 
& 0.2964 & 0.7256 & 0.6120 & 0.9209 & 0.1147 & 40.13
& 0.6861 & 0.4788
& 2.928
\\
nano-banana 
& 0.1189 & 0.8724 & 0.8093 & 0.9584 & 0.0440 & 72.30
& 0.8070 & 0.6331 
& 6.600
\\
\rowcolor{gray!10} Seedream4.0 
& 0.1671 & 0.8787 & 0.7878 & 0.9613 & 0.0486 & 66.66
& 0.7197 & 0.5612 
& 6.719
\\
Qwen-Image-Edit-Plus 
& 0.1916 & 0.8705 & 0.7833 & 0.9416 & 0.0535 & 58.00
& 0.6807 & 0.5456  
& 5.753
\\
\rowcolor{gray!10} Sora 
& 0.1736 & 0.8753 & 0.7143 & 0.9524 & 0.0686 & 62.21
& 0.4328 & 0.3758  
& 5.536
\\
\hline
\rowcolor[HTML]{D7F6FF}
Ours 
& \textbf{0.0918} & \textbf{0.9457} & \textbf{0.8919} & \textbf{0.9765} & \textbf{0.0164} & \textbf{81.91}
& \textbf{0.8573} & \textbf{0.7569}
& \textbf{6.747}
\end{tabular}}}
\end{table*}

\begin{table*}[t]\small
\centering
\scriptsize{
\captionsetup{font=small}
\caption{Ablation study results across all metrics indicate the necessity of each step in our approach.}
\label{tab:ablation}
\vspace{-10pt}
\resizebox{\linewidth}{!}{
\setlength\tabcolsep{3.pt}
\renewcommand\arraystretch{1.1}
\begin{tabular}{r||cccccc|cc|c}
\hline
\hline
\rowcolor{gray!20}
 & 
\multicolumn{6}{c|}{\textbf{Overall Local Refine}} & 
\multicolumn{2}{c|}{\textbf{Face Local Refine}} &
\multicolumn{1}{c}{\textbf{Text Local Refine}}
\\
\cline{2-10} 
\rowcolor{gray!20}
\multirow{-2}{*}{Methods}  
& Dreamsim$\downarrow$ & DINOv2$\uparrow$ & DINOv3$\uparrow$ & Clip-I$\uparrow$ & MSE$\downarrow$ & VLM Score$\uparrow$
& ArcFace$\uparrow$ & LVFace$\uparrow$
& VLM Text Identify$\uparrow$
\\
\hline\hline
Ours(w/o RL) 
& 0.0929 & 0.9214 & 0.8653 & 0.9701 & 0.0250 & 74.01
& 0.8265 & 0.7152  
& 5.286
\\
\rowcolor{gray!10} Ours(EasyControl PE\&w/o RL) 
& 0.1016 & 0.9126 & 0.8614 & 0.9646 & 0.0266 & 72.15
& 0.8163 & 0.7060 
& 5.157
\\
Ours(RL Qwen-Image-Edit-Plus) 
& 0.1449 & 0.8846 & 0.8585 & 0.9583 & 0.0298 & 68.52
& 0.7481 & 0.6452
& 6.443
\\
\rowcolor{gray!10} Ours(w/o masked MSELoss) 
& \textbf{0.0916} & 0.9160 & 0.8864 & 0.9637 & 0.0177 & 74.68
& 0.8507 & 0.7496 
& 5.911
\\
\hline
\rowcolor[HTML]{D7F6FF}
Ours(full) 
& 0.0918 & \textbf{0.9457} & \textbf{0.8919} & \textbf{0.9765} & \textbf{0.0164} & \textbf{81.91}
& \textbf{0.8573} & \textbf{0.7569}
& \textbf{6.747}
\\
\vspace{-10pt}
\end{tabular}}}
\end{table*}

We utilize Flux.1-Kontext-dev~\cite{labs2025flux1kontextflowmatching} as our base model and adopt a LoRA strategy (rank 128) on a single H200 GPU with a resolution of $512 \times 512$, a batch size of 1, and a learning rate of $1 \times 10^{-4}$, followed by a reinforcement learning phase on four H200 GPUs using GRPO with 16 trajectories, a batch size of 4, and 800 fine-tuning steps at the same learning rate.

\noindent \textbf{Baseline Methods.}  
Our comparison covers both open-source and closed-source state-of-the-art multi-image input models, including gemini-2.5-flash-image-preview (nano-banana)~\cite{nanobanana}, Seedream4.0~\cite{seedream2025seedream}, Sora Image~\cite{liu2024sora}, Qwen-Image-Edit-Plus~\cite{wu2025qwen}, Mosaic~\cite{she2025mosaicmultisubjectpersonalizedgeneration}, and DreamO~\cite{mou2025dreamo}.

\noindent \textbf{Metrics.}  
For consistency evaluation, we compute the following similarity metrics: CLIP Image Score~\cite{radford2021learning}, DreamSim~\cite{fu2023dreamsim}, DINOv2~\cite{oquab2023dinov2}, and DINOv3~\cite{simeoni2025dinov3} image similarity scores, Mean Squared Error, and the advanced Vision-Language Models (VLMs) gemini2.5~\cite{comanici2025gemini} Score. For facial similarity evaluation, we adopt recent open-source state-of-the-art methods such as LVFace~\cite{you2025lvface}, as well as the widely used ArcFace~\cite{deng2019arcface} similarity metric. For text restoration tasks, since current open-source methods are insufficient for recognizing fine details and small characters in real-world scenarios, we carefully design prompts that enable gemini2.5 to perform cross-image text comparison and scoring. 

\noindent \textbf{Benchmarks.}  
For evaluation, we introduce a benchmark named Detail400, which includes branded products with text and logos, clothing with printed patterns, vehicles, and jewelry or decorative items. These categories are known to frequently suffer from detail preservation and transfer failure in generative models, making them ideal for assessing fine-grained detail consistency.

\subsection{Comparison and Evaluation}

\noindent \textbf{Qualitative Evaluation.}
\begin{figure*}[tbp]
  \centering
  \includegraphics[width=1\textwidth]{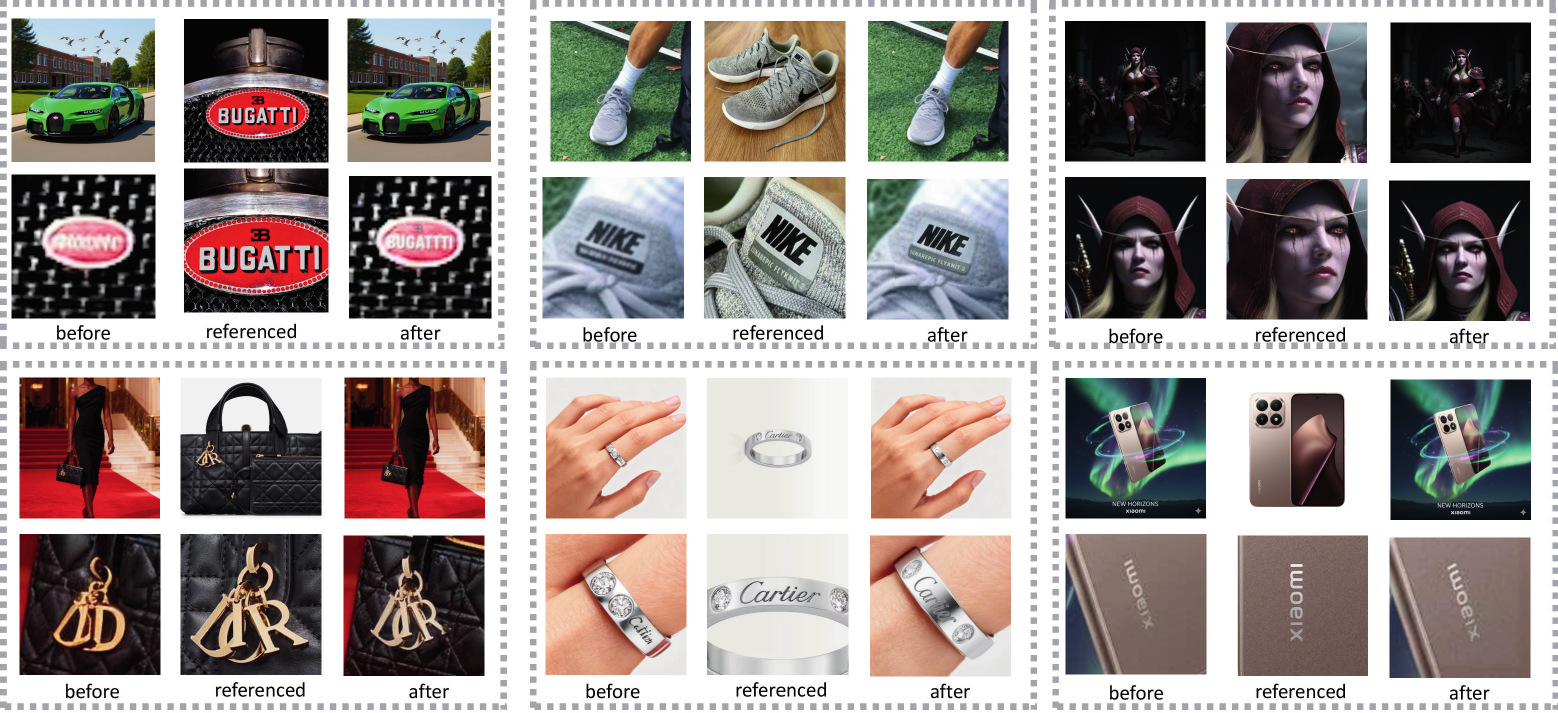}
  \caption{Qualitative results demonstrate that our method can accurately restore fine details in images.}
  \label{fig:method_results}
\end{figure*}
As shown in Fig.~\ref{fig:method_comp}, our method outperforms existing state-of-the-art approaches in both text and image reference-based restoration. In contrast, current methods often fail to faithfully reconstruct the image according to the reference: some redraw the image based on their own interpretation rather than the reference, while others overuse or even directly copy-paste the reference image, resulting in incorrect lighting, structure, and perspective of fine details. Fig.~\ref{fig:method_results} presents additional restoration examples, showing that our method performs well in detail restoration tasks. It not only faithfully reconstructs according to the reference image but also preserves the original lighting and structural integrity.

\noindent \textbf{Quantitative Evaluation.} 
Table \ref{tab:compare_sota_vqa} and \ref{tab:ablation} present the quantitative evaluation results. Experiments show that our generated results outperform both open-source and closed-source state-of-the-art models across multiple similarity metrics. Moreover, our method achieves superior facial detail restoration and demonstrates leading performance in text reconstruction.

\subsection{Ablation Study.}
\begin{figure}[tbp]
  \centering
  \includegraphics[width=0.45\textwidth]{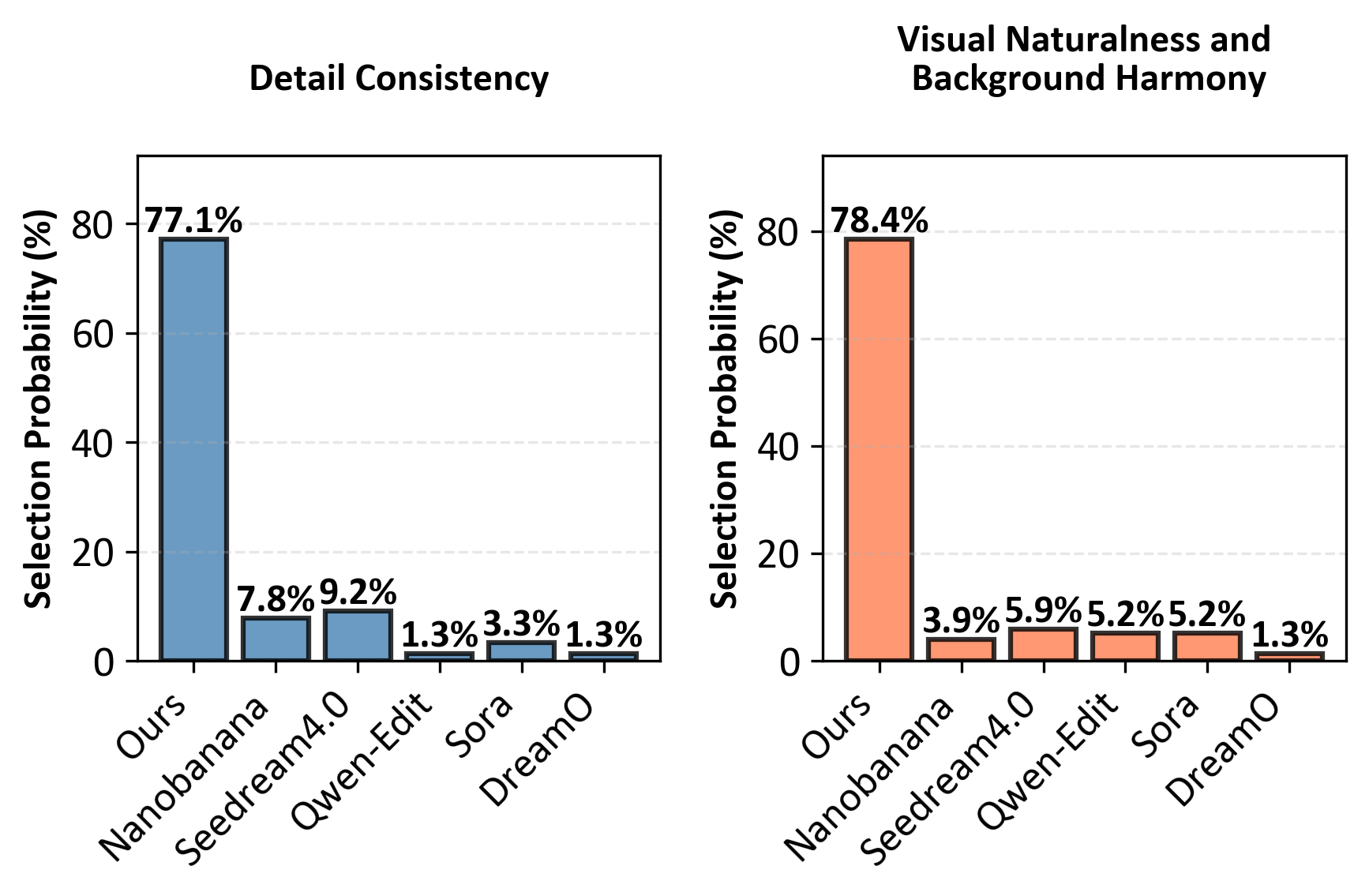}
  \caption{The user study demonstrates that our method achieves the highest human preference alignment in both detail consistency and restoration naturalness.}
  \label{fig:user_study}
  \vspace{-16pt}
\end{figure}

\begin{figure}[h]
  \centering
  \includegraphics[width=\linewidth]{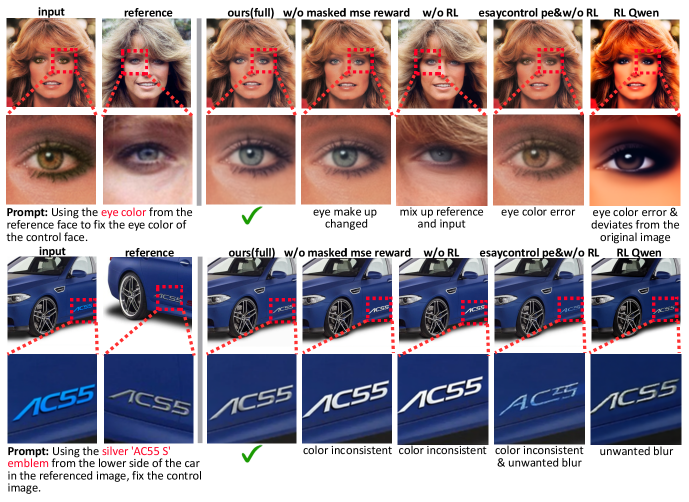}
  \caption{Ablation studies demonstrate that each component of our method (GRPO, position embedding, and masked MSE reward) is essential. Even when applying GRPO fine-tuning to Qwen-Image-Edit-Plus, a model that natively supports multi-image inputs, using exactly the same data, the results still fall short of ours.}
  \label{fig:ablation}
  \vspace{-16pt}
\end{figure}

Our ablation study aims to verify the following key components of our method: (1) Our modification to the position embedding scheme outperforms common control-input designs used in existing works under our research problem settings, such as the ID clone approach adopted by Easy Control; (2) The reinforcement learning stage is indispensable for achieving the final model performance; (3) Our design surpasses the results of directly post-training recent open-source state-of-the-art models with native multi-image input support, such as Qwen-Image-Edit-Plus; and (4) The Masked MSE Score in our reward function plays a crucial role, as it directly determines whether the model can effectively learn the corresponding fine details. Table.~\ref{tab:ablation} presents the quantitative results of the ablation study, and Fig.~\ref{fig:ablation} shows the corresponding visual results. Together, these findings demonstrate that the position embedding modification, GRPO training, and Masked MSE Score design in our method are all indispensable components.

\subsection{User Study.}

To further validate the effectiveness of our method in real-world perceptual settings, we conducted a user study. We designed an online survey and collected responses from 17 participants with experience in image editing, design, or AI media tools. In each question, participants were shown results generated by different methods and were asked to select the generated result that best matched the local details and the one that appeared most natural and seamless after local restoration. The aggregated user preference results are shown in Fig.~\ref{fig:user_study}. Statistical results show that our method not only achieves the highest perceived consistency with the reference image in terms of fine-grained details, but also produces the most natural and seamless results. In contrast, other methods either fail to preserve local detail consistency with the reference or introduce noticeable stitching artifacts after editing. These results confirm that the improvements of the refined image achieved by OmniRefiner are not only measurable by quantitative metrics, but also clearly recognized by human observers.

\section{Limitation and Future Work}
Currently, the selection of regions requiring refinement and the retrieval of suitable reference patches are performed manually. This human involvement limits full automation and scalability. In future work, we plan to incorporate vision–language models and retrieval-augmented generation (RAG) to automatically identify refinement regions and retrieve optimal reference details, enabling an end-to-end autonomous refinement pipeline.

\section{Conclusion}

We propose \textbf{OmniRefiner}, a unified framework for reference-guided fine detail enhancement that serves as a plug-and-play refinement module for modern diffusion models. By extending single-image diffusion Transformers into dual-input generators through simple yet effective Position Embedding modifications, and employing a two-stage learning paradigm, namely supervised fine-tuning for task adaptation followed by reinforcement learning with specially designed reward functions that promote both global similarity and local detail fidelity. We further introduce a scalable automated quadruplet data pipeline that enables high-quality training without manual annotation. Extensive experiments demonstrate that OmniRefiner achieves superior detail recovery across diverse content and architectures, validating the importance of explicit reference conditioning and reward-driven refinement. We believe this work provides a principled pathway toward high-fidelity, controllable, and reliable image refinement, and hope it inspires future research in fine-grained detail generation modeling and reference-aligned visual synthesis.

{
    \small
    \bibliographystyle{ieeenat_fullname}
    \bibliography{main}

\begin{thebibliography}{68}
\providecommand{\natexlab}[1]{#1}
\providecommand{\url}[1]{\texttt{#1}}
\expandafter\ifx\csname urlstyle\endcsname\relax
  \providecommand{\doi}[1]{doi: #1}\else
  \providecommand{\doi}{doi: \begingroup \urlstyle{rm}\Url}\fi

\bibitem[Black et~al.(2023)Black, Janner, Du, Kostrikov, and Levine]{Black2023TrainingDM}
Kevin Black, Michael Janner, Yilun Du, Ilya Kostrikov, and Sergey Levine.
\newblock Training diffusion models with reinforcement learning.
\newblock \emph{ArXiv}, abs/2305.13301, 2023.

\bibitem[Brack et~al.(2023)Brack, Friedrich, Kornmeier, Tsaban, Schramowski, Kersting, and Passos]{Brack2023LEDITSLI}
Manuel Brack, Felix Friedrich, Katharina Kornmeier, Linoy Tsaban, Patrick Schramowski, Kristian Kersting, and Apolin'ario Passos.
\newblock Ledits++: Limitless image editing using text-to-image models.
\newblock \emph{2024 IEEE/CVF Conference on Computer Vision and Pattern Recognition (CVPR)}, pages 8861--8870, 2023.

\bibitem[Cao et~al.(2023)Cao, Wang, Qi, Shan, Qie, and Zheng]{Cao_2023_ICCV}
Mingdeng Cao, Xintao Wang, Zhongang Qi, Ying Shan, Xiaohu Qie, and Yinqiang Zheng.
\newblock Masactrl: Tuning-free mutual self-attention control for consistent image synthesis and editing.
\newblock In \emph{Proceedings of the IEEE/CVF International Conference on Computer Vision (ICCV)}, pages 22560--22570, 2023.

\bibitem[Chen et~al.(2025{\natexlab{a}})Chen, Zhao, Sun, Chen, Wang, Du, and Wu]{chen2025xverseconsistentmultisubjectcontrol}
Bowen Chen, Mengyi Zhao, Haomiao Sun, Li Chen, Xu Wang, Kang Du, and Xinglong Wu.
\newblock Xverse: Consistent multi-subject control of identity and semantic attributes via dit modulation, 2025{\natexlab{a}}.

\bibitem[Chen et~al.(2024)Chen, Huang, Liu, Shen, Zhao, and Zhao]{Chen_2024_CVPR}
Xi Chen, Lianghua Huang, Yu Liu, Yujun Shen, Deli Zhao, and Hengshuang Zhao.
\newblock Anydoor: Zero-shot object-level image customization.
\newblock In \emph{Proceedings of the IEEE/CVF Conference on Computer Vision and Pattern Recognition (CVPR)}, pages 6593--6602, 2024.

\bibitem[Chen et~al.(2025{\natexlab{b}})Chen, Chen, and Song]{chen2025transanimate}
Xuewei Chen, Zhimin Chen, and Yiren Song.
\newblock Transanimate: Taming layer diffusion to generate rgba video.
\newblock \emph{arXiv preprint arXiv:2503.17934}, 2025{\natexlab{b}}.

\bibitem[Comanici et~al.(2025)Comanici, Bieber, Schaekermann, Pasupat, Sachdeva, Dhillon, Blistein, Ram, Zhang, Rosen, et~al.]{comanici2025gemini}
Gheorghe Comanici, Eric Bieber, Mike Schaekermann, Ice Pasupat, Noveen Sachdeva, Inderjit Dhillon, Marcel Blistein, Ori Ram, Dan Zhang, Evan Rosen, et~al.
\newblock Gemini 2.5: Pushing the frontier with advanced reasoning, multimodality, long context, and next generation agentic capabilities.
\newblock \emph{arXiv preprint arXiv:2507.06261}, 2025.

\bibitem[Deng et~al.(2025)Deng, Zhu, Li, Gou, Li, Wang, Zhong, Yu, Nie, Song, et~al.]{deng2025emerging}
Chaorui Deng, Deyao Zhu, Kunchang Li, Chenhui Gou, Feng Li, Zeyu Wang, Shu Zhong, Weihao Yu, Xiaonan Nie, Ziang Song, et~al.
\newblock Emerging properties in unified multimodal pretraining.
\newblock \emph{arXiv preprint arXiv:2505.14683}, 2025.

\bibitem[Deng et~al.(2019)Deng, Guo, Xue, and Zafeiriou]{deng2019arcface}
Jiankang Deng, Jia Guo, Niannan Xue, and Stefanos Zafeiriou.
\newblock Arcface: Additive angular margin loss for deep face recognition.
\newblock In \emph{Proceedings of the IEEE/CVF conference on computer vision and pattern recognition}, pages 4690--4699, 2019.

\bibitem[Deutch et~al.(2024)Deutch, Gal, Garibi, Patashnik, and Cohen-Or]{Deutch2024TurboEditTI}
Gilad Deutch, Rinon Gal, Daniel Garibi, Or Patashnik, and Daniel Cohen-Or.
\newblock Turboedit: Text-based image editing using few-step diffusion models.
\newblock \emph{SIGGRAPH Asia 2024 Conference Papers}, 2024.

\bibitem[Esser et~al.(2024)Esser, Kulal, Blattmann, Entezari, M{\"u}ller, Saini, Levi, Lorenz, Sauer, Boesel, et~al.]{esser2024scaling}
Patrick Esser, Sumith Kulal, Andreas Blattmann, Rahim Entezari, Jonas M{\"u}ller, Harry Saini, Yam Levi, Dominik Lorenz, Axel Sauer, Frederic Boesel, et~al.
\newblock Scaling rectified flow transformers for high-resolution image synthesis.
\newblock In \emph{Forty-first international conference on machine learning}, 2024.

\bibitem[Fortin et~al.(2025)Fortin, Vernade, Kampf, and Reshi]{nanobanana}
Alisa Fortin, Guillaume Vernade, Kat Kampf, and Ammaar Reshi.
\newblock Introducing gemini 2.5 flash image, our state-of-the-art image model.
\newblock https://developers.googleblog.com/en/introducing-gemini-2-5-flash-image/, 2025.

\bibitem[Fu et~al.(2023)Fu, Tamir, Sundaram, Chai, Zhang, Dekel, and Isola]{fu2023dreamsim}
Stephanie Fu, Netanel Tamir, Shobhita Sundaram, Lucy Chai, Richard Zhang, Tali Dekel, and Phillip Isola.
\newblock Dreamsim: Learning new dimensions of human visual similarity using synthetic data.
\newblock \emph{arXiv preprint arXiv:2306.09344}, 2023.

\bibitem[Gong et~al.(2025)Gong, Song, Li, Li, and Zhang]{gong2025relationadapter}
Yan Gong, Yiren Song, Yicheng Li, Chenglin Li, and Yin Zhang.
\newblock Relationadapter: Learning and transferring visual relation with diffusion transformers.
\newblock \emph{arXiv preprint arXiv:2506.02528}, 2025.

\bibitem[Guo et~al.(2025)Guo, Zeng, Song, Zhang, Zhang, and Liu]{guo2025any2anytryon}
Hailong Guo, Bohan Zeng, Yiren Song, Wentao Zhang, Chuang Zhang, and Jiaming Liu.
\newblock Any2anytryon: Leveraging adaptive position embeddings for versatile virtual clothing tasks.
\newblock \emph{arXiv preprint arXiv:2501.15891}, 2025.

\bibitem[Hertz et~al.(2022)Hertz, Mokady, Tenenbaum, Aberman, Pritch, and Cohen-Or]{Hertz2022PrompttoPromptIE}
Amir Hertz, Ron Mokady, Jay~M. Tenenbaum, Kfir Aberman, Yael Pritch, and Daniel Cohen-Or.
\newblock Prompt-to-prompt image editing with cross attention control.
\newblock \emph{ArXiv}, abs/2208.01626, 2022.

\bibitem[Ho et~al.(2020)Ho, Jain, and Abbeel]{ho2020denoising}
Jonathan Ho, Ajay Jain, and Pieter Abbeel.
\newblock Denoising diffusion probabilistic models.
\newblock \emph{Advances in neural information processing systems}, 33:\penalty0 6840--6851, 2020.

\bibitem[Huang et~al.(2025)Huang, Song, Zhang, Guo, Wang, and Liu]{huang2025arteditor}
Shijie Huang, Yiren Song, Yuxuan Zhang, Hailong Guo, Xueyin Wang, and Jiaming Liu.
\newblock Arteditor: Learning customized instructional image editor from few-shot examples.
\newblock In \emph{Proceedings of the IEEE/CVF International Conference on Computer Vision}, pages 17651--17662, 2025.

\bibitem[Jiang et~al.(2025)Jiang, Gu, Song, Tsang, and Shou]{jiang2025personalized}
Yuxin Jiang, Yuchao Gu, Yiren Song, Ivor Tsang, and Mike~Zheng Shou.
\newblock Personalized vision via visual in-context learning.
\newblock \emph{arXiv preprint arXiv:2509.25172}, 2025.

\bibitem[Kotar et~al.(2023)Kotar, Tian, Yu, Yamins, and Wu]{kotar2023these}
Klemen Kotar, Stephen Tian, Hong-Xing Yu, Dan Yamins, and Jiajun Wu.
\newblock Are these the same apple? comparing images based on object intrinsics.
\newblock \emph{Advances in Neural Information Processing Systems}, 36:\penalty0 40853--40871, 2023.

\bibitem[Kulikov et~al.(2024)Kulikov, Kleiner, Huberman-Spiegelglas, and Michaeli]{Kulikov2024FlowEditIT}
Vladimir Kulikov, Matan Kleiner, Inbar Huberman-Spiegelglas, and Tomer Michaeli.
\newblock Flowedit: Inversion-free text-based editing using pre-trained flow models.
\newblock \emph{ArXiv}, abs/2412.08629, 2024.

\bibitem[Labs(2024)]{flux2024}
Black~Forest Labs.
\newblock Flux.
\newblock \url{https://github.com/black-forest-labs/flux}, 2024.

\bibitem[Labs et~al.(2025)Labs, Batifol, Blattmann, Boesel, Consul, Diagne, Dockhorn, English, English, Esser, Kulal, Lacey, Levi, Li, Lorenz, Müller, Podell, Rombach, Saini, Sauer, and Smith]{labs2025flux1kontextflowmatching}
Black~Forest Labs, Stephen Batifol, Andreas Blattmann, Frederic Boesel, Saksham Consul, Cyril Diagne, Tim Dockhorn, Jack English, Zion English, Patrick Esser, Sumith Kulal, Kyle Lacey, Yam Levi, Cheng Li, Dominik Lorenz, Jonas Müller, Dustin Podell, Robin Rombach, Harry Saini, Axel Sauer, and Luke Smith.
\newblock Flux.1 kontext: Flow matching for in-context image generation and editing in latent space, 2025.

\bibitem[Li et~al.(2024)Li, Cao, Wang, Qi, Cheng, and Shan]{Li_2024_CVPR}
Zhen Li, Mingdeng Cao, Xintao Wang, Zhongang Qi, Ming-Ming Cheng, and Ying Shan.
\newblock Photomaker: Customizing realistic human photos via stacked id embedding.
\newblock In \emph{Proceedings of the IEEE/CVF Conference on Computer Vision and Pattern Recognition (CVPR)}, pages 8640--8650, 2024.

\bibitem[Lin et~al.(2025)Lin, Li, Cheng, Niu, Ye, He, Yuan, Yu, Wang, Ge, et~al.]{lin2025uniworld}
Bin Lin, Zongjian Li, Xinhua Cheng, Yuwei Niu, Yang Ye, Xianyi He, Shenghai Yuan, Wangbo Yu, Shaodong Wang, Yunyang Ge, et~al.
\newblock Uniworld: High-resolution semantic encoders for unified visual understanding and generation.
\newblock \emph{arXiv preprint arXiv:2506.03147}, 2025.

\bibitem[Lipman et~al.(2022)Lipman, Chen, Ben-Hamu, Nickel, and Le]{lipman2022flow}
Yaron Lipman, Ricky~TQ Chen, Heli Ben-Hamu, Maximilian Nickel, and Matt Le.
\newblock Flow matching for generative modeling.
\newblock \emph{arXiv preprint arXiv:2210.02747}, 2022.

\bibitem[Liu et~al.(2025)Liu, Liu, Liang, Li, Liu, Wang, Wan, Zhang, and Ouyang]{liu2025flow}
Jie Liu, Gongye Liu, Jiajun Liang, Yangguang Li, Jiaheng Liu, Xintao Wang, Pengfei Wan, Di Zhang, and Wanli Ouyang.
\newblock Flow-grpo: Training flow matching models via online rl.
\newblock \emph{arXiv preprint arXiv:2505.05470}, 2025.

\bibitem[Liu et~al.(2024)Liu, Zhang, Li, Yan, Gao, Chen, Yuan, Huang, Sun, Gao, et~al.]{liu2024sora}
Yixin Liu, Kai Zhang, Yuan Li, Zhiling Yan, Chujie Gao, Ruoxi Chen, Zhengqing Yuan, Yue Huang, Hanchi Sun, Jianfeng Gao, et~al.
\newblock Sora: A review on background, technology, limitations, and opportunities of large vision models.
\newblock \emph{arXiv preprint arXiv:2402.17177}, 2024.

\bibitem[Lu et~al.(2025)Lu, Zhang, Liu, Wang, and Song]{lu2025easytext}
Runnan Lu, Yuxuan Zhang, Jiaming Liu, Haofan Wang, and Yiren Song.
\newblock Easytext: Controllable diffusion transformer for multilingual text rendering.
\newblock \emph{arXiv preprint arXiv:2505.24417}, 2025.

\bibitem[Luo et~al.(2023)Luo, Tan, Huang, Li, and Zhao]{luo2023latent}
Simian Luo, Yiqin Tan, Longbo Huang, Jian Li, and Hang Zhao.
\newblock Latent consistency models: Synthesizing high-resolution images with few-step inference.
\newblock \emph{arXiv preprint arXiv:2310.04378}, 2023.

\bibitem[Ma et~al.(2024)Ma, Liu, Wang, Pan, He, Yuan, Zeng, Cai, Shum, Liu, et~al.]{ma2024followyouremoji}
Yue Ma, Hongyu Liu, Hongfa Wang, Heng Pan, Yingqing He, Junkun Yuan, Ailing Zeng, Chengfei Cai, Heung-Yeung Shum, Wei Liu, et~al.
\newblock Follow-your-emoji: Fine-controllable and expressive freestyle portrait animation.
\newblock In \emph{SIGGRAPH Asia 2024 Conference Papers}, pages 1--12, 2024.

\bibitem[Ma et~al.(2025{\natexlab{a}})Ma, Feng, Hu, Wang, Wang, Zheng, He, Zhu, Liu, He, et~al.]{ma2025controllable}
Yue Ma, Kunyu Feng, Zhongyuan Hu, Xinyu Wang, Yucheng Wang, Mingzhe Zheng, Xuanhua He, Chenyang Zhu, Hongyu Liu, Yingqing He, et~al.
\newblock Controllable video generation: A survey.
\newblock \emph{arXiv preprint arXiv:2507.16869}, 2025{\natexlab{a}}.

\bibitem[Ma et~al.(2025{\natexlab{b}})Ma, Feng, Zhang, Liu, Zhang, Xing, Zhang, Yang, Wang, and Chen]{ma2025followcreation}
Yue Ma, Kunyu Feng, Xinhua Zhang, Hongyu Liu, David~Junhao Zhang, Jinbo Xing, Yinhan Zhang, Ayden Yang, Zeyu Wang, and Qifeng Chen.
\newblock Follow-your-creation: Empowering 4d creation through video inpainting.
\newblock \emph{arXiv preprint arXiv:2506.04590}, 2025{\natexlab{b}}.

\bibitem[Ma et~al.(2025{\natexlab{c}})Ma, He, Wang, Wang, Shen, Qi, Ying, Cai, Li, Shum, et~al.]{ma2025followyourclick}
Yue Ma, Yingqing He, Hongfa Wang, Andong Wang, Leqi Shen, Chenyang Qi, Jixuan Ying, Chengfei Cai, Zhifeng Li, Heung-Yeung Shum, et~al.
\newblock Follow-your-click: Open-domain regional image animation via motion prompts.
\newblock In \emph{Proceedings of the AAAI Conference on Artificial Intelligence}, pages 6018--6026, 2025{\natexlab{c}}.

\bibitem[Ma et~al.(2025{\natexlab{d}})Ma, Liu, Zhu, Yang, Feng, Zhang, Li, Han, Qi, and Chen]{ma2025followyourmotion}
Yue Ma, Yulong Liu, Qiyuan Zhu, Ayden Yang, Kunyu Feng, Xinhua Zhang, Zhifeng Li, Sirui Han, Chenyang Qi, and Qifeng Chen.
\newblock Follow-your-motion: Video motion transfer via efficient spatial-temporal decoupled finetuning.
\newblock \emph{arXiv preprint arXiv:2506.05207}, 2025{\natexlab{d}}.

\bibitem[Ma et~al.(2025{\natexlab{e}})Ma, Yan, Liu, Wang, Pan, He, Yuan, Zeng, Cai, Shum, et~al.]{ma2025followfaster}
Yue Ma, Zexuan Yan, Hongyu Liu, Hongfa Wang, Heng Pan, Yingqing He, Junkun Yuan, Ailing Zeng, Chengfei Cai, Heung-Yeung Shum, et~al.
\newblock Follow-your-emoji-faster: Towards efficient, fine-controllable, and expressive freestyle portrait animation.
\newblock \emph{arXiv preprint arXiv:2509.16630}, 2025{\natexlab{e}}.

\bibitem[Mao et~al.(2025)Mao, Zhang, Pan, Jiang, Han, Liu, and Zhou]{mao2025ace++}
Chaojie Mao, Jingfeng Zhang, Yulin Pan, Zeyinzi Jiang, Zhen Han, Yu Liu, and Jingren Zhou.
\newblock Ace++: Instruction-based image creation and editing via context-aware content filling.
\newblock \emph{arXiv preprint arXiv:2501.02487}, 2025.

\bibitem[Mou et~al.(2024)Mou, Wang, Xie, Wu, Zhang, Qi, and Shan]{mou2024t2i}
Chong Mou, Xintao Wang, Liangbin Xie, Yanze Wu, Jian Zhang, Zhongang Qi, and Ying Shan.
\newblock T2i-adapter: Learning adapters to dig out more controllable ability for text-to-image diffusion models.
\newblock In \emph{Proceedings of the AAAI conference on artificial intelligence}, pages 4296--4304, 2024.

\bibitem[Mou et~al.(2025)Mou, Wu, Wu, Guo, Zhang, Cheng, Luo, Ding, Zhang, Li, et~al.]{mou2025dreamo}
Chong Mou, Yanze Wu, Wenxu Wu, Zinan Guo, Pengze Zhang, Yufeng Cheng, Yiming Luo, Fei Ding, Shiwen Zhang, Xinghui Li, et~al.
\newblock Dreamo: A unified framework for image customization.
\newblock \emph{arXiv preprint arXiv:2504.16915}, 2025.

\bibitem[Oquab et~al.(2023)Oquab, Darcet, Moutakanni, Vo, Szafraniec, Khalidov, Fernandez, Haziza, Massa, El-Nouby, et~al.]{oquab2023dinov2}
Maxime Oquab, Timoth{\'e}e Darcet, Th{\'e}o Moutakanni, Huy Vo, Marc Szafraniec, Vasil Khalidov, Pierre Fernandez, Daniel Haziza, Francisco Massa, Alaaeldin El-Nouby, et~al.
\newblock Dinov2: Learning robust visual features without supervision.
\newblock \emph{arXiv preprint arXiv:2304.07193}, 2023.

\bibitem[Peebles and Xie(2023{\natexlab{a}})]{dit}
William Peebles and Saining Xie.
\newblock Scalable diffusion models with transformers.
\newblock In \emph{Proceedings of the IEEE/CVF international conference on computer vision}, pages 4195--4205, 2023{\natexlab{a}}.

\bibitem[Peebles and Xie(2023{\natexlab{b}})]{peebles2023scalable}
William Peebles and Saining Xie.
\newblock Scalable diffusion models with transformers.
\newblock In \emph{Proceedings of the IEEE/CVF international conference on computer vision}, pages 4195--4205, 2023{\natexlab{b}}.

\bibitem[Podell et~al.(2023)Podell, English, Lacey, Blattmann, Dockhorn, M{\"u}ller, Penna, and Rombach]{podell2023sdxl}
Dustin Podell, Zion English, Kyle Lacey, Andreas Blattmann, Tim Dockhorn, Jonas M{\"u}ller, Joe Penna, and Robin Rombach.
\newblock Sdxl: Improving latent diffusion models for high-resolution image synthesis.
\newblock \emph{arXiv preprint arXiv:2307.01952}, 2023.

\bibitem[Radford et~al.(2021)Radford, Kim, Hallacy, Ramesh, Goh, Agarwal, Sastry, Askell, Mishkin, Clark, et~al.]{radford2021learning}
Alec Radford, Jong~Wook Kim, Chris Hallacy, Aditya Ramesh, Gabriel Goh, Sandhini Agarwal, Girish Sastry, Amanda Askell, Pamela Mishkin, Jack Clark, et~al.
\newblock Learning transferable visual models from natural language supervision.
\newblock In \emph{International conference on machine learning}, pages 8748--8763. PmLR, 2021.

\bibitem[Rombach et~al.(2022)Rombach, Blattmann, Lorenz, Esser, and Ommer]{rombach2022high}
Robin Rombach, Andreas Blattmann, Dominik Lorenz, Patrick Esser, and Bj{\"o}rn Ommer.
\newblock High-resolution image synthesis with latent diffusion models.
\newblock In \emph{Proceedings of the IEEE/CVF conference on computer vision and pattern recognition}, pages 10684--10695, 2022.

\bibitem[Ronneberger et~al.(2015)Ronneberger, Fischer, and Brox]{ronneberger2015u}
Olaf Ronneberger, Philipp Fischer, and Thomas Brox.
\newblock U-net: Convolutional networks for biomedical image segmentation.
\newblock In \emph{International Conference on Medical image computing and computer-assisted intervention}, pages 234--241. Springer, 2015.

\bibitem[Seedream et~al.(2025)Seedream, Chen, Gao, Gong, Guo, Guo, Guo, Hou, Huang, Huang, et~al.]{seedream2025seedream}
Team Seedream, Yunpeng Chen, Yu Gao, Lixue Gong, Meng Guo, Qiushan Guo, Zhiyao Guo, Xiaoxia Hou, Weilin Huang, Yixuan Huang, et~al.
\newblock Seedream 4.0: Toward next-generation multimodal image generation.
\newblock \emph{arXiv preprint arXiv:2509.20427}, 2025.

\bibitem[Shao et~al.(2024)Shao, Wang, Zhu, Xu, Song, Bi, Zhang, Zhang, Li, Wu, et~al.]{shao2024deepseekmath}
Zhihong Shao, Peiyi Wang, Qihao Zhu, Runxin Xu, Junxiao Song, Xiao Bi, Haowei Zhang, Mingchuan Zhang, YK Li, Yang Wu, et~al.
\newblock Deepseekmath: Pushing the limits of mathematical reasoning in open language models.
\newblock \emph{arXiv preprint arXiv:2402.03300}, 2024.

\bibitem[She et~al.(2025)She, Fu, Liu, Jin, Wang, Liu, and Jiang]{she2025mosaicmultisubjectpersonalizedgeneration}
Dong She, Siming Fu, Mushui Liu, Qiaoqiao Jin, Hualiang Wang, Mu Liu, and Jidong Jiang.
\newblock Mosaic: Multi-subject personalized generation via correspondence-aware alignment and disentanglement, 2025.

\bibitem[Shi et~al.(2024)Shi, Song, Zhang, Liu, and Zou]{shi2024fonts}
Wenda Shi, Yiren Song, Dengming Zhang, Jiaming Liu, and Xingxing Zou.
\newblock Fonts: Text rendering with typography and style controls.
\newblock \emph{arXiv preprint arXiv:2412.00136}, 2024.

\bibitem[Shi et~al.(2025)Shi, Song, Rao, Zhang, Liu, and Zou]{shi2025wordcon}
Wenda Shi, Yiren Song, Zihan Rao, Dengming Zhang, Jiaming Liu, and Xingxing Zou.
\newblock Wordcon: Word-level typography control in scene text rendering.
\newblock \emph{arXiv preprint arXiv:2506.21276}, 2025.

\bibitem[Sim{\'e}oni et~al.(2025)Sim{\'e}oni, Vo, Seitzer, Baldassarre, Oquab, Jose, Khalidov, Szafraniec, Yi, Ramamonjisoa, et~al.]{simeoni2025dinov3}
Oriane Sim{\'e}oni, Huy~V Vo, Maximilian Seitzer, Federico Baldassarre, Maxime Oquab, Cijo Jose, Vasil Khalidov, Marc Szafraniec, Seungeun Yi, Micha{\"e}l Ramamonjisoa, et~al.
\newblock Dinov3.
\newblock \emph{arXiv preprint arXiv:2508.10104}, 2025.

\bibitem[Song et~al.(2020)Song, Meng, and Ermon]{song2020denoising}
Jiaming Song, Chenlin Meng, and Stefano Ermon.
\newblock Denoising diffusion implicit models.
\newblock \emph{arXiv preprint arXiv:2010.02502}, 2020.

\bibitem[Song et~al.(2024)Song, Huang, Yao, Ye, Ci, Liu, Zhang, and Shou]{song2024processpainter}
Yiren Song, Shijie Huang, Chen Yao, Xiaojun Ye, Hai Ci, Jiaming Liu, Yuxuan Zhang, and Mike~Zheng Shou.
\newblock Processpainter: Learn painting process from sequence data.
\newblock \emph{arXiv preprint arXiv:2406.06062}, 2024.

\bibitem[Song et~al.(2025{\natexlab{a}})Song, Chen, and Shou]{song2025layertracer}
Yiren Song, Danze Chen, and Mike~Zheng Shou.
\newblock Layertracer: Cognitive-aligned layered svg synthesis via diffusion transformer.
\newblock \emph{arXiv preprint arXiv:2502.01105}, 2025{\natexlab{a}}.

\bibitem[Song et~al.(2025{\natexlab{b}})Song, Liu, and Shou]{song2025makeanything}
Yiren Song, Cheng Liu, and Mike~Zheng Shou.
\newblock Makeanything: Harnessing diffusion transformers for multi-domain procedural sequence generation.
\newblock \emph{arXiv preprint arXiv:2502.01572}, 2025{\natexlab{b}}.

\bibitem[Song et~al.(2025{\natexlab{c}})Song, Liu, and Shou]{song2025omniconsistency}
Yiren Song, Cheng Liu, and Mike~Zheng Shou.
\newblock Omniconsistency: Learning style-agnostic consistency from paired stylization data.
\newblock \emph{arXiv preprint arXiv:2505.18445}, 2025{\natexlab{c}}.

\bibitem[Wallace et~al.(2024)Wallace, Dang, Rafailov, Zhou, Lou, Purushwalkam, Ermon, Xiong, Joty, and Naik]{wallace2024diffusion}
Bram Wallace, Meihua Dang, Rafael Rafailov, Linqi Zhou, Aaron Lou, Senthil Purushwalkam, Stefano Ermon, Caiming Xiong, Shafiq Joty, and Nikhil Naik.
\newblock Diffusion model alignment using direct preference optimization.
\newblock In \emph{Proceedings of the IEEE/CVF Conference on Computer Vision and Pattern Recognition}, pages 8228--8238, 2024.

\bibitem[Wang et~al.(2025)Wang, Zhao, Zhou, Lu, Li, and Song]{wang2025diffdecompose}
Zitong Wang, Hang Zhao, Qianyu Zhou, Xuequan Lu, Xiangtai Li, and Yiren Song.
\newblock Diffdecompose: Layer-wise decomposition of alpha-composited images via diffusion transformers.
\newblock \emph{arXiv preprint arXiv:2505.21541}, 2025.

\bibitem[Wu et~al.(2025)Wu, Li, Zhou, Lin, Gao, Yan, Yin, Bai, Xu, Chen, et~al.]{wu2025qwen}
Chenfei Wu, Jiahao Li, Jingren Zhou, Junyang Lin, Kaiyuan Gao, Kun Yan, Sheng-ming Yin, Shuai Bai, Xiao Xu, Yilei Chen, et~al.
\newblock Qwen-image technical report.
\newblock \emph{arXiv preprint arXiv:2508.02324}, 2025.

\bibitem[Xiao et~al.(2025)Xiao, Wang, Zhou, Yuan, Xing, Yan, Li, Wang, Huang, and Liu]{xiao2025omnigen}
Shitao Xiao, Yueze Wang, Junjie Zhou, Huaying Yuan, Xingrun Xing, Ruiran Yan, Chaofan Li, Shuting Wang, Tiejun Huang, and Zheng Liu.
\newblock Omnigen: Unified image generation.
\newblock In \emph{Proceedings of the Computer Vision and Pattern Recognition Conference}, pages 13294--13304, 2025.

\bibitem[Xue et~al.(2025)Xue, Wu, Gao, Kong, Zhu, Chen, Liu, Liu, Guo, Huang, et~al.]{xue2025dancegrpo}
Zeyue Xue, Jie Wu, Yu Gao, Fangyuan Kong, Lingting Zhu, Mengzhao Chen, Zhiheng Liu, Wei Liu, Qiushan Guo, Weilin Huang, et~al.
\newblock Dancegrpo: Unleashing grpo on visual generation.
\newblock \emph{arXiv preprint arXiv:2505.07818}, 2025.

\bibitem[Yang et~al.(2023)Yang, Tao, Lyu, Ge, Chen, Li, Shen, Zhu, and Li]{Yang2023UsingHF}
Kai Yang, Jian Tao, Jiafei Lyu, Chunjiang Ge, Jiaxin Chen, Qimai Li, Weihan Shen, Xiaolong Zhu, and Xiu Li.
\newblock Using human feedback to fine-tune diffusion models without any reward model.
\newblock \emph{2024 IEEE/CVF Conference on Computer Vision and Pattern Recognition (CVPR)}, pages 8941--8951, 2023.

\bibitem[You et~al.(2025)You, Li, Sun, Wei, Guo, Feng, and Ran]{you2025lvface}
Jinghan You, Shanglin Li, Yuanrui Sun, Jiangchuan Wei, Mingyu Guo, Chao Feng, and Jiao Ran.
\newblock Lvface: Progressive cluster optimization for large vision models in face recognition.
\newblock \emph{arXiv preprint arXiv:2501.13420}, 2025.

\bibitem[Zhang et~al.(2023)Zhang, Rao, and Agrawala]{zhang2023adding}
Lvmin Zhang, Anyi Rao, and Maneesh Agrawala.
\newblock Adding conditional control to text-to-image diffusion models.
\newblock In \emph{Proceedings of the IEEE/CVF international conference on computer vision}, pages 3836--3847, 2023.

\bibitem[Zhang et~al.(2024)Zhang, Wei, Zhang, Song, Liu, Li, Tang, Hu, and Zhao]{zhang2024stable}
Yuxuan Zhang, Lifu Wei, Qing Zhang, Yiren Song, Jiaming Liu, Huaxia Li, Xu Tang, Yao Hu, and Haibo Zhao.
\newblock Stable-makeup: When real-world makeup transfer meets diffusion model.
\newblock \emph{arXiv preprint arXiv:2403.07764}, 2024.

\bibitem[Zhang et~al.(2025{\natexlab{a}})Zhang, Yuan, Song, Wang, and Liu]{zhang2025easycontrol}
Yuxuan Zhang, Yirui Yuan, Yiren Song, Haofan Wang, and Jiaming Liu.
\newblock Easycontrol: Adding efficient and flexible control for diffusion transformer.
\newblock \emph{arXiv preprint arXiv:2503.07027}, 2025{\natexlab{a}}.

\bibitem[Zhang et~al.(2025{\natexlab{b}})Zhang, Zhang, Song, Zhang, Tang, and Liu]{zhang2025stable}
Yuxuan Zhang, Qing Zhang, Yiren Song, Jichao Zhang, Hao Tang, and Jiaming Liu.
\newblock Stable-hair: Real-world hair transfer via diffusion model.
\newblock In \emph{Proceedings of the AAAI Conference on Artificial Intelligence}, pages 10348--10356, 2025{\natexlab{b}}.

\end{thebibliography}
}

\appendix

\onecolumn
\begin{center}
    \Large
    \textbf{\thetitle}\\
    \vspace{0.5em}Supplementary Material \\
    \vspace{0.65em}
\end{center}

\noindent Our supplementary material is organized as follows: Section~\ref{sec:user_study} provides the details of our user study; Section~\ref{sec:eval-detail} describes our evaluation protocol, including how we employed and configured VLMs for assessment, and Section~\ref{sec:visual-result} presents additional results and comparisons, including those from our ablation study;

\section{User Study Detail}\label{sec:user_study}

\begin{figure*}[h]
  \centering
  \includegraphics[width=1\textwidth]{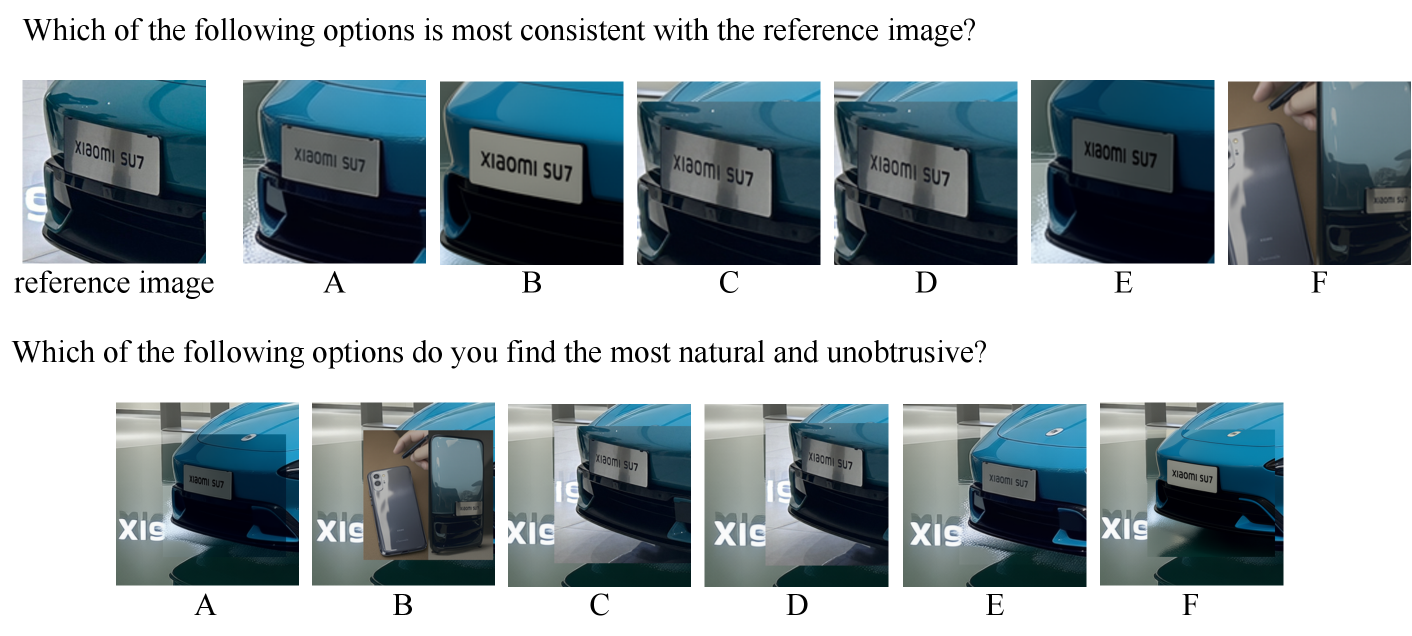}
  \caption{User study example: participants were first informed that detail consistency must be achieved while maintaining natural integration with the background, and they were also asked to directly select the result that appeared most natural and unobtrusive.}
  \label{fig:sup_user_study}
\end{figure*}

Our user study involved 17 participants, all of whom had prior experience with image editing tools or AI-based image/video generation models. For each question in the survey, participants were shown randomly ordered results generated by six different methods, including ours. To avoid bias, the names of all methods were anonymized and the order was fully randomized.
To prevent misunderstandings regarding the evaluation objective, we explicitly clarified to participants that superior detail consistency does not simply mean copying the reference image onto the target; rather, it requires preserving fine details while ensuring seamless integration with the target image. Additionally, to ensure that participants had ample opportunity to express which method produced the most natural restoration in their view, we added an extra question for each test case that asked them to select the result that appeared the most natural and unobtrusive.
This design allowed participants to choose the image that best aligned with their perception of visual quality without repeatedly comparing how closely each method matched the reference details. As a result, the scientific rigor of our comparative evaluation is further ensured.

\section{Evaluation Detail}\label{sec:eval-detail}

\begin{tcolorbox}[
  colback=gray!5,    
  colframe=gray!40,  
  boxrule=0.5pt,     
  arc=2mm,           
  left=4pt, right=4pt, top=4pt, bottom=4pt, 
  title=Evaluation Prompt for Overall Local Refine,
  label=prop:prompt_1
]
You are an expert-level VLM evaluator tasked with scoring image generation models.
Your objective is to provide a precise, quantitative assessment of a Generated Image by comparing it to a Ground Truth (GT) Image. You will be given both images and must return a score out of 100 points based on the specific rubric provided below.

\textbf{INPUT:}
\begin{itemize}
    \item Image 1: [Generated Image]
    \item Image 2: [Ground Truth (GT) Image]
\end{itemize}

\textbf{SCORING RUBRIC (100 POINTS TOTAL)}

\textbf{1. Overall Consistency (40 Points)}
\begin{itemize}
    \item \textbf{Description:} Evaluate the macro-level alignment between the Generated Image and the GT Image. This includes:
    \begin{itemize}
        \item Composition: Are objects and subjects placed in the correct locations?
        \item Color Palette: Do the overall colors and tones match?
        \item Lighting \& Shadows: Is the direction, hardness, and intensity of light and shadows consistent with the GT?
    \end{itemize}
    \item \textbf{Score:} Assign a score from 0 to 40 based on how closely the generated image matches the GT in these three aspects.
\end{itemize}

\textbf{2. Detail Consistency (40 Points)}
\begin{itemize}
    \item \textbf{Description:} Evaluate the micro-level, high-frequency detail fidelity. The Generated Image must faithfully reproduce specific elements from the GT. Pay close attention to:
    \begin{itemize}
        \item Text \& Typography: Is all text identical, legible, and correctly rendered?
        \item Logos \& Insignia: Are brand marks, logos, or symbols accurately replicated?
        \item Fine Patterns \& Textures: Are intricate details (e.g., fabric weave, wood grain, complex patterns) preserved?
        \item Facial Details: Are facial features, identity, and expression consistent?
    \end{itemize}
    \item \textbf{Score:} Assign a score from 0 to 40. High scores require near-perfect replication of all specified details. Deduct points heavily for errors in text, logos, or faces.
\end{itemize}

\textbf{3. Generated Image Quality (20 Points)}
\begin{itemize}
    \item \textbf{Description:} Evaluate the intrinsic quality of the Generated Image on its own, independent of the GT. The image should be clear, sharp, and free from common generation issues.
    \item \textbf{Criteria:}
    \begin{itemize}
        \item Clarity \& Sharpness: Is the image sharp, or is it blurry, pixelated, or soft?
        \item Artifacts: Is the image clean, or does it contain visual artifacts (e.g., `waxy' skin, malformed limbs, strange distortions, digital noise, color banding)?
    \end{itemize}
    \item \textbf{Score:} Assign a score from 0 to 20 based on its technical quality and realism. A flawless, high-resolution image gets 20. An image with significant artifacts or blurriness gets a low score.
\end{itemize}
\textbf{OUTPUT FORMAT}

You MUST provide your evaluation in the following strict format:

\begin{verbatim}
{
    "Thinking": <Your detailed reasoning process here>,
    "Overall Consistency": <Score out of 40>,
    "Detail Consistency": <Score out of 40>,
    "Generated Image Quality": <Score out of 20>,
    "Total Score": <Total Score out of 100>
}
\end{verbatim}
\end{tcolorbox}

In our evaluation, DreamO, Xverse, and Qwen-Image-Edit-Plus, together with our OmniRefiner, were executed on a single H200 GPU, with the generation seed fixed to 42. For Seedream4.0, Nanobanana, and Sora, we used their officially provided API endpoints for generation. Notably, Sora refused to produce outputs for approximately 20\% of the test cases—mainly those involving face or logo restoration. For such cases, we retried up to five times; ultimately, about 15\% of the cases still failed to generate any output, and these were excluded from Sora’s results. All generated images were kept at the same resolution as the input, with side lengths ranging from 768 to 2048. For VLM-based evaluation, we used the Gemini 2.5 Pro model in non-inference mode, and the image and text evaluation prompts used for scoring are shown in~\ref{prop:prompt_1}.

\begin{tcolorbox}[
  colback=gray!5,    
  colframe=gray!40,  
  boxrule=0.5pt,     
  arc=2mm,           
  left=4pt, right=4pt, top=4pt, bottom=4pt, 
  title=Evaluation Prompt for Text Rendering Accuracy,
  label=prop:prompt_2
]
\footnotesize
\textbf{Role:} You are a high-precision VLM (Visual Language Model) evaluator. Your specialization is analyzing and scoring the fidelity of text rendering in generated images.

\textbf{Task:} You will be given two images:

[Image 1]: A ``Generated Image'' produced by a model.

[Image 2]: A ``Ground Truth Image'' (the real picture).

Your task is to compare [Image 1] to [Image 2] and provide a score from 0 to 10 that evaluates only the accuracy of the rendered text. You must follow the steps and scoring rubric below.

\textbf{Instructions}

\textbf{Step 1: Analysis} Before scoring, you must first perform and mentally note the following analysis:

\begin{itemize}
    \item \textbf{Identify Text Regions (Image 1):} Scan the ``Generated Image'' and identify all areas that contain text or text-like artifacts.
    \item \textbf{Identify Text Regions (Image 2):} Scan the ``Ground Truth Image'' and identify all areas that contain text.
    \item \textbf{Evaluate Region Correspondence:} Compare your findings. Does the text in [Image 1] appear on the correct objects and surfaces as seen in [Image 2]?
    \item \textbf{Evaluate Content \& Detail Accuracy:} Compare the content of the text. Is the text in [Image 1] readable, correctly spelled, and do the details (e.g., small text on labels, distant signs) match [Image 2]?
\end{itemize}

\textbf{Step 2: Scoring}

After your analysis, assign a single score based strictly on the following rubric.

\begin{itemize}
    \item \textbf{8-10 points:}
    \begin{itemize}
        \item The generated text is completely accurate (correct spelling, no gibberish).
        \item The text fully corresponds to the ground truth (correct location, correct text).
        \item The text in all detail regions (small print, complex text) is also well-reconstructed.
    \end{itemize}
    
    \item \textbf{5-8 points:}
    \begin{itemize}
        \item The generated text is meaningful and largely correct.
        \item The text regions are correspondent (text is in the right places).
        \item However, the text in some detail regions is incorrect, missing, or blurry.
    \end{itemize}
    
    \item \textbf{2-5 points:}
    \begin{itemize}
        \item The generated text is meaningful (e.g., recognizable words) and may have only a few artifacts.
        \item But, the text regions are incorrect (e.g., text is on the wrong object, or major text is missing).
    \end{itemize}
    
    \item \textbf{0-2 points:}
    \begin{itemize}
        \item The generated text is not correct text.
        \item It consists of meaningless, text-like artifacts or ``gibberish'' that only mimics the shape of text.
    \end{itemize}
\end{itemize}

\textbf{Step 3: Output}

Provide your evaluation in the following exact format. Do not add any conversational text outside of this structure.

\begin{verbatim}
[ANALYSIS]
<Briefly describe where you see text/artifacts in the generated image.>
<Briefly describe where you see text in the ground truth image.>
<Your assessment of location matching, e.g., "Good", "Poor", "Perfect".>
<Your assessment of text content, e.g., "Perfectly accurate", 
"Mostly correct but fails on details", "Completely incorrect artifacts".>

[SCORING RATIONALE]
<Provide a brief justification for your score, linking your analysis 
to the scoring rubric. For example: "Score is 7 because regions correspond, 
but text on the small bottle is incorrect.">

[FINAL SCORE]
<A single number between 0 and 10>
\end{verbatim}
\end{tcolorbox}

\section{Visual Results}\label{sec:visual-result}

\subsection{More Visual Results}

\begin{figure*}[h]
  \centering
  \includegraphics[width=0.9\textwidth]{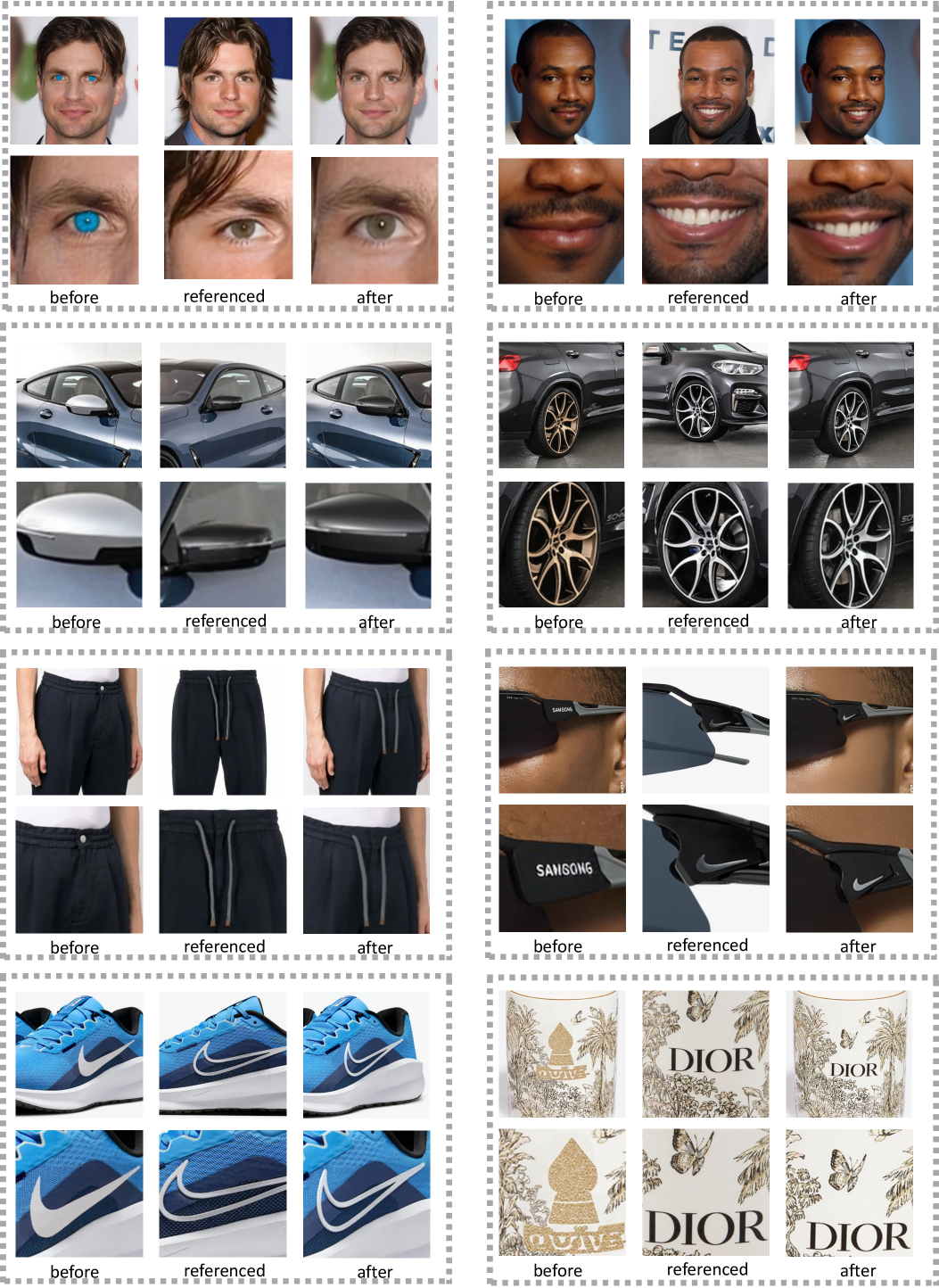}
  \caption{Additional restoration results show that our method is highly effective in both detail-guided refinement and detail-guided replacement.}
  \label{fig:sup_results}
\end{figure*}

\subsection{More Visual Results of Method Comparison}

\begin{figure*}[h]
  \centering
  \includegraphics[width=1\textwidth]{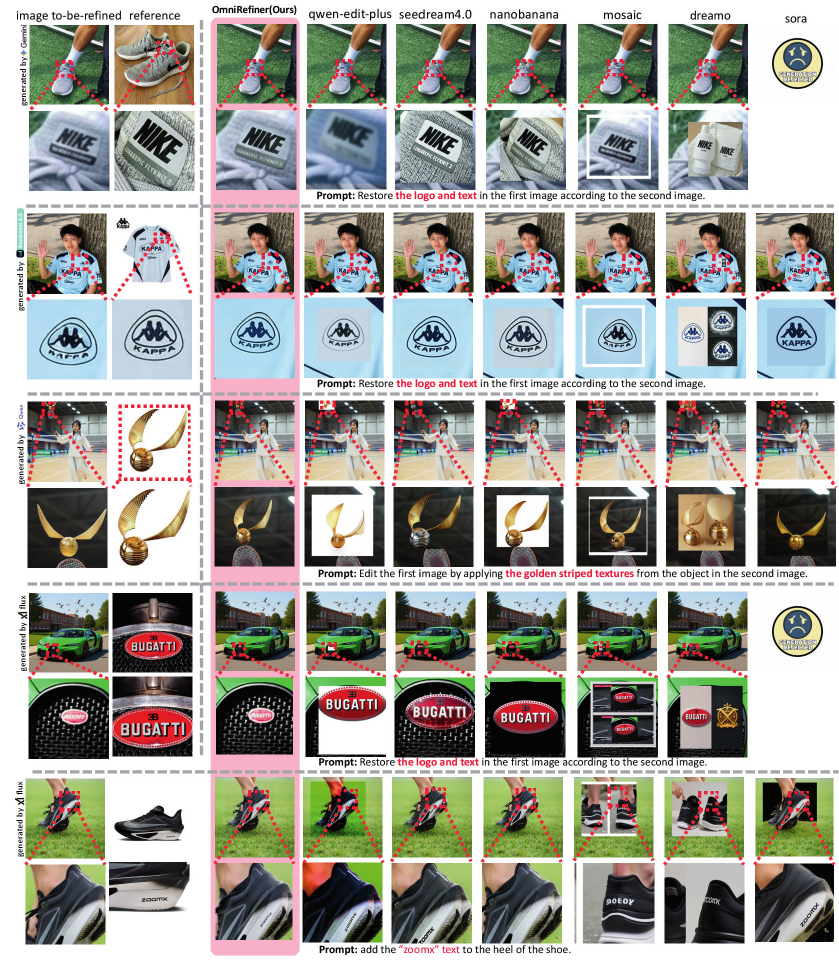}
  \caption{Object-centric method comparison results reveal that our approach exhibits advantages in the coherence of text, logos, and textures, as well as the quality and consistency of the overall repair results.}
  \label{fig:sup_objects}
\end{figure*}

\begin{figure*}[h]
  \centering
  \includegraphics[width=1\textwidth]{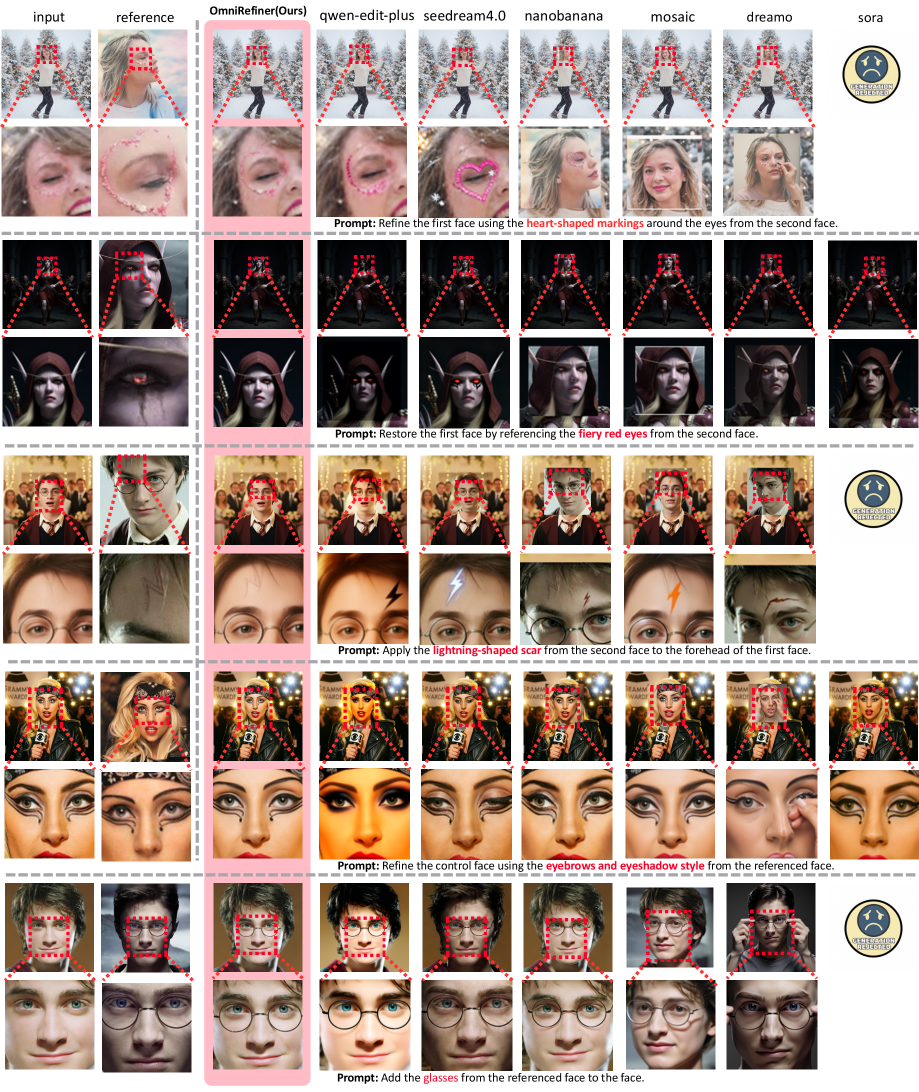}
  \caption{Facial comparison results further demonstrate that our method achieves superior understanding of human faces, yielding outstanding restoration quality for fine details such as eyeshadow, eyebrows, scars, and iris color.}
  \label{fig:sup_faces}
\end{figure*}

\subsection{More Visual Results of Ablation Study}

\begin{figure*}[h]
  \centering
  \includegraphics[width=0.85\textwidth]{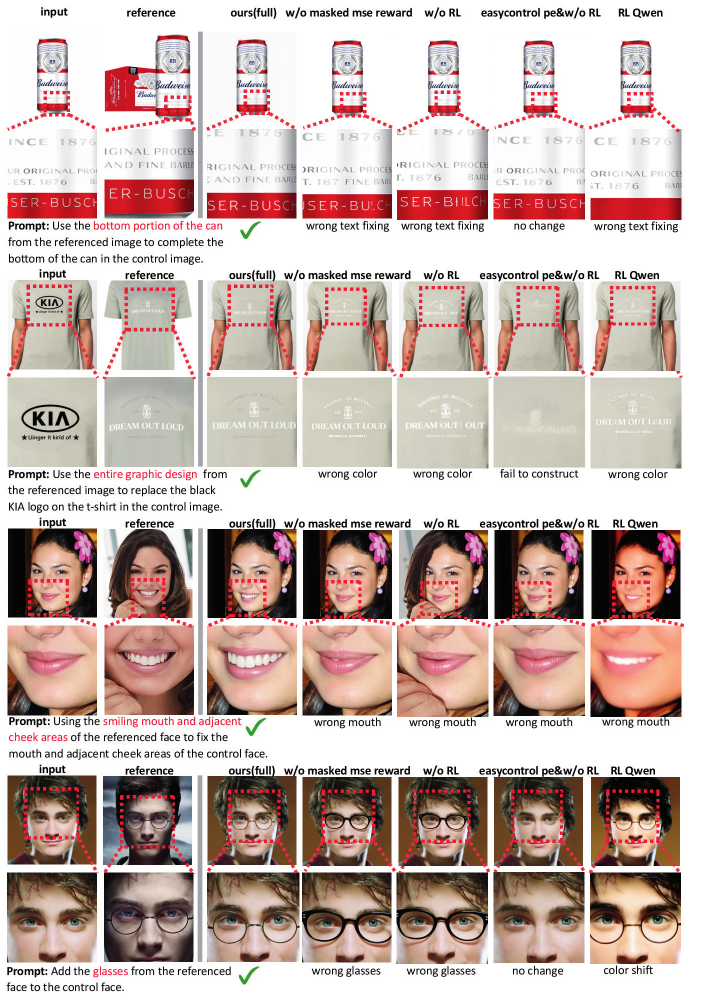}
  \caption{These further ablation results collectively demonstrate the crucial importance of our design in enhancing the model's understanding, repair, reconstruction, and replacement of details.}
  \label{fig:sup_ablation}
\end{figure*}



\end{document}